\documentclass[10pt,twocolumn,letterpaper]{article}

\usepackage{iccv}              
\definecolor{iccvblue}{rgb}{0.21,0.49,0.74}
\usepackage[pagebackref,breaklinks,colorlinks,allcolors=iccvblue]{hyperref}

\def\paperID{*****} 
\def\confName{ICCV}
\def\confYear{2025}

\title{Deterministic Object Pose Conﬁdence Region Estimation}

\author{
    Jinghao Wang$^{1}$\quad Zhang Li$^{1}$\quad Zi Wang$^{1}$\thanks{Corresponding author. 
    {\tt\small wangzi16@nudt.edu.cn} }\quad Banglei Guan$^{1}$\quad Yang Shang$^{1}$\quad Qifeng Yu$^{1}$ \\
   $^{1}$ National University of Defense Technology}

\begin{document}
\maketitle
\begin{abstract}
6D pose confidence region estimation has emerged as a critical direction, aiming to perform uncertainty quantification for assessing the reliability of estimated poses. 
However, current sampling-based approach suffers from critical limitations that severely impede their practical deployment:
1) the sampling speed significantly decreases as the number of samples increases. 
2) the derived confidence regions are often excessively large. 
To address these challenges, we propose a deterministic and efficient method for estimating pose confidence regions. 
Our approach uses inductive conformal prediction to calibrate the deterministically regressed Gaussian keypoint distributions into 2D keypoint confidence regions. 
We then leverage the implicit function theorem to propagate these keypoint confidence regions directly into 6D pose confidence regions. 
This method avoids the inefficiency and inflated region sizes associated with sampling and ensembling.
It provides compact confidence regions that cover the ground-truth poses with a user-defined confidence level.
Experimental results on the LineMOD Occlusion and SPEED datasets show that our method achieves higher pose estimation accuracy with reduced computational time. 
For the same coverage rate, our method yields significantly smaller confidence region volumes, reducing them by up to 99.9\% for rotations and 99.8\% for translations.
The code will be available soon.
\end{abstract}    
\section{Introduction}
\label{sec:intro}

Determining the 6D pose of an object from an RGB image is a fundamental task in computer vision, 
with extensive applications in autonomous driving~\cite{Yurtsever2020survey}, 
robotic manipulation~\cite{manuelli2019kpam}, augmented reality~\cite{su2019deep}, 
and space robotics~\cite{chen2019satellite}. 
The majority of pose estimation research~\cite{wu2018real, kehl2017ssd, xiang2018posecnn, hsiao2024confronting, yang2023object, schmeckpeper2022semantic, peng2019pvnet} primarily focuses on delivering a single optimal pose estimation. 
It is often adequate for applications that tolerate occasional inaccuracies. 
However, it underperforms where high reliability and precise uncertainty quantification (UQ) are essential. 
Because visual ambiguities undermine the accuracy of even the best algorithms,
 leading to disasters in safety-critical scenarios.

\begin{figure}[t]
    \centering
    \begin{subfigure}[t]{\columnwidth}
        \includegraphics[width=\columnwidth]{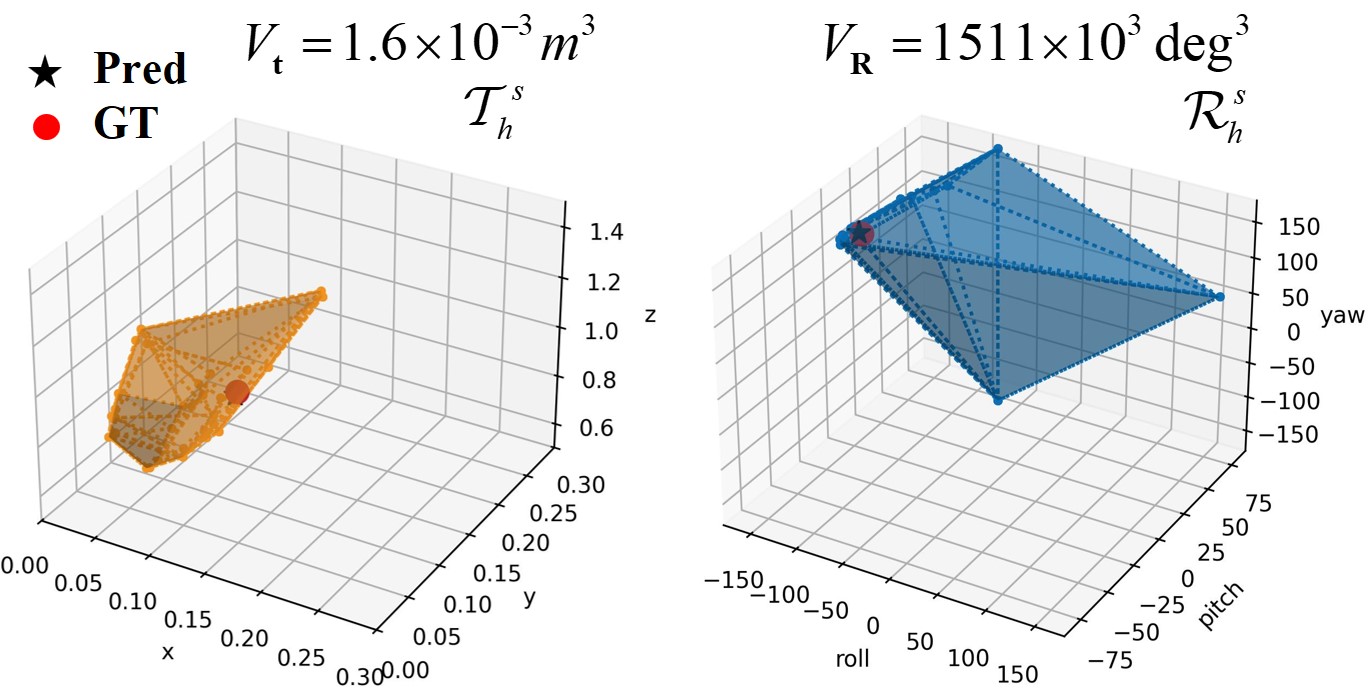}  
        \caption{Sampling method~\cite{yang2023object}}
        \label{fig:top1}
    \end{subfigure}
    
    \vspace{0.5em}  
    
    \begin{subfigure}[t]{\columnwidth}
        \includegraphics[width=\columnwidth]{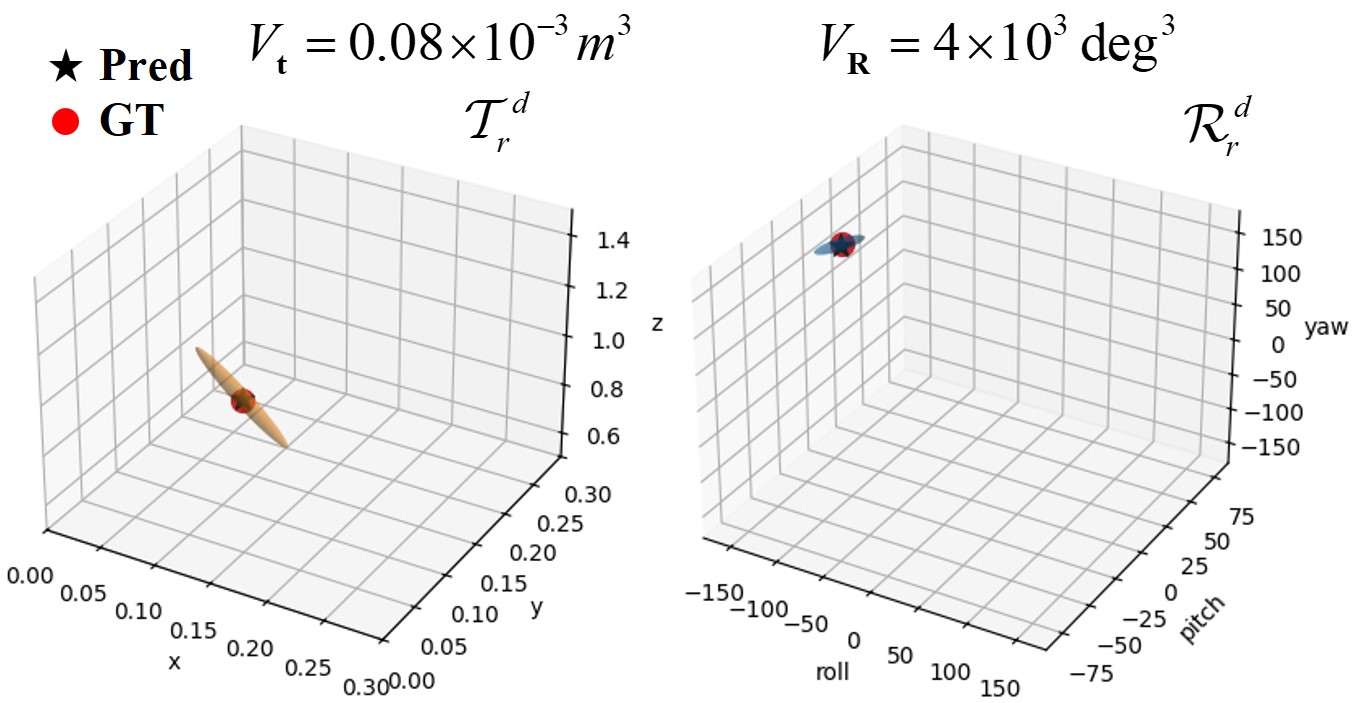}  
        \caption{Deterministic method}
        \label{fig:top2}
    \end{subfigure}
    
    \caption{
    Narrower confidence regions of our deterministic approach compared to the sampling method~\cite{yang2023object}. (a) A heatmap-guided keypoints are established for 2D confidence regions $\mathcal{K}^{h}$ via ICP, with 6D pose confidence regions 
    $\mathcal{T}^{s}_h$ and $\mathcal{R}^{s}_h$ inferred through sampling. 
    (b) We directly regress keypoint Gaussian distributions and generated $\mathcal{K}^{r}$ by ICP. 
    To avoid inefficiency and broadened regions due to sampling, 
    we apply a Jacobian matrix from the IFT to propagate  $\mathcal{K}^{r}$  into the 6D pose.
    }
    \label{fig:structure_top}
\end{figure}

Driven by the need for safety and reliability in real-world applications, 
UQ has been a critical research focus in robotics for decades. 
UQ is vital for ensuring successful object grasping and prevents collisions, 
and it can be categorized as either a probabilistic distribution of outcomes or a scalar confidence~\cite{shi2021fast}.
However, both distribution estimation~\cite{Gilitschenski2020,okorn2020learning,deng2022deep_bingham,sato2023Probabilistic} and confidence estimation~\cite{wang2019DenseFusion,tremblay2018corl} are not always reliable and often fail to provide statistically guaranteed confidence regions. 
Additionally, evaluating predictive uncertainty, especially in 6D pose estimation, is inherently challenging due to the lack of ground truth uncertainty.

A practical approach involves deriving 6D pose prediction confidence regions,
which are designed to cover the ground truth poses with a user-specified probability. 
As depicted in Fig.~\ref{fig:structure_top}, 
the method proposed by Yang et al.~\cite{yang2023object} 
uses inductive conformal prediction (ICP)~\cite{Papadopoulos2008InductiveCP} to conformalize heatmaps into keypoint confidence regions  $\mathcal{K}^{h}$. 
ICP provides distribution-free, finite-sample guarantees by constructing confidence regions that adapt based on the nonconformity between predictions and observations.
By applying ICP to keypoint detection, 
we can specify a threshold to generate a confidence region that covers the true position with a certain probability.
Subsequently,
a sampling-based method constructs 6D pose confidence regions $\mathcal{C}^{s}_h=\{\mathcal{R}^{s}_h,\mathcal{T}^{s}_h\}$, 
consisting of confidence regions for rotation ($\mathcal{R}^{s}_h$) and translation ($\mathcal{T}^{s}_h$). 
For the confidence region, 
the superscripts $s$ denote the sampling approach, 
while the subscripts $h$ represent the heatmap method. 
Although $\mathcal{C}^{s}_h$  can cover the ground truth, 
the sampling process increases the confidence region volume and reduces computational efficiency.
Furthermore, common metrics for pose estimation, 
such as reprojection, 5$^{\circ}$/5cm,  ADD(-S)~\cite{pvnet_pami}, VSD, MSSD, and MSPD~\cite{hodan2018bop}, 

assess correctness based on whether errors fall below specific thresholds.
These metrics focus solely on measurement accuracy but do not account for the estimation regions' size.

To address these issues, 
we present a deterministic method for estimating pose confidence regions as shown in~\cref{fig:structure_top}.
We incorporate uncertainty into the loss function to efficiently regress keypoint Gaussian distributions. 
Utilizing ICP, 
these distributions are calibrated into keypoint confidence regions $\mathcal{K}^{r}$.
To avoid expanding the confidence region due to low-quality sampled poses, 
we deterministically propagate \(\mathcal{K}^{r}\) into the pose confidence region \(\mathcal{C}^{d}_{r}\) using the Jacobian derived from the implicit function theorem (IFT).
The superscripts $d$ denote the deterministical approach, 
while the subscripts $r$ represent the regress method. 
Additionally, we introduce metrics for evaluating measurement uncertainty, 
including coverage rates and radii for 2D keypoint confidence regions, 
as well as the coverage rates and volumes of 6D pose confidence region.
These metrics provide a more comprehensive assessment of the estimated pose. 
On the SPEED~\cite{Kisantal2019Satellite} and LMO~\cite{lmo_dataset_BrachmannKMGSR14} datasets, our method reduces the pose confidence region volume by 63.8\%, 99.9\% for rotations and 92.2\%, 99.8\% for translations.
In summary, our work makes the following contributions:
\begin{enumerate}
    \item We present an ICP-based method for predicting keypoint confidence regions, which relies on deterministic regression of Gaussian keypoint distributions. 
    \item Leveraging IFT, we propagate the keypoint confidence regions directly into the 6D pose, 
    while maintaining a pose coverage rate comparable to that of~\cite{yang2023object}.
    \item We propose  thorough metrics to evaluate the estimated pose confidence region.

\end{enumerate}
\begin{figure*}[t]
    \centering
    \includegraphics[width=0.99\textwidth]{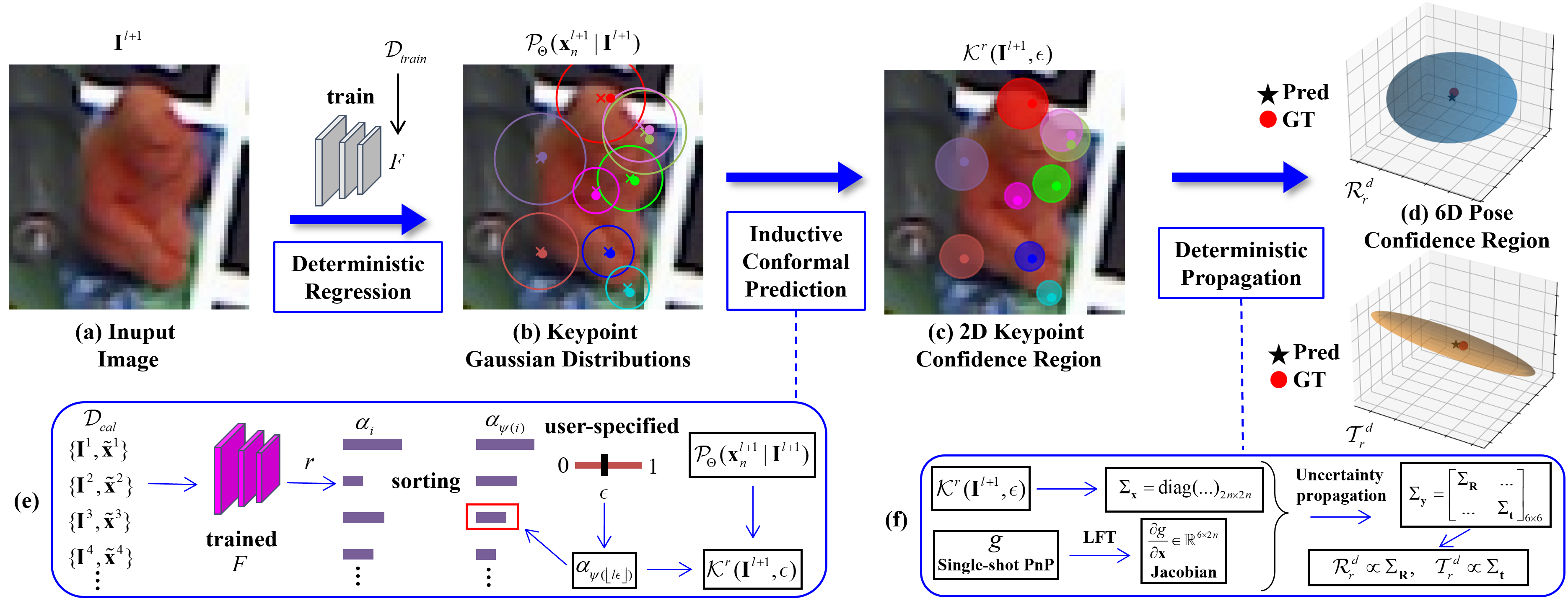}  
    \caption{
Given an input image, we deterministically regress semantic keypoints (crosses) along with their uncertainties (hollow circles) to generate 2D keypoint confidence regions (translucent circles). 
These regions are calibrated to ensure coverage of the ground truth keypoints (solid dots), achieving, e.g., 90\% coverage using inductive conformal prediction. 
We then apply the PnP algorithm once to predict the 6D pose (pentagram). 
Through IFT, we propagate 2D into 6D confidence regions around the predicted pose, maintaining the same coverage rate for the ground truth (red dots).
    }
    \label{fig_structure}
\end{figure*}
\section{Related Work}
\subsection{Single Point Object Pose Estimation} 

Current pose estimation methods can be categorized into regression-based and correspondence-based approaches. 
Regression-based methods directly recover object poses from images, either in a coupled~\cite{Manhardt2019Explaining, kleeberger2020single, Bengtson2021IROS} or decoupled manner~\cite{wu2018real, kehl2017ssd, xiang2018posecnn, hsiao2024confronting}. 
They show better computational efficiency but still struggle with the nonlinear nature of rotation representation. 
Correspondence-based methods establish dense~\cite{li2019cdpn,hodan2020epos} or sparse correspondences~\cite{pavlakos20176ICRA, Rad2017ICCVBB} between the input image and a 3D model, then use algorithms like PnP~\cite{fischler1981random} for pose estimation. Dense methods offer higher accuracy but at the cost of significant computational overhead, while sparse methods predict predefined keypoints via voting~\cite{peng2019pvnet}, heatmaps~\cite{schmeckpeper2022semantic,yang2023object}, or direct regression~\cite{Rad2017ICCVBB, tekin2018real}. 
Recent approaches integrate differentiable PnP~\cite{brachmann2018learning, chen2020end, iwase2021repose} into end-to-end pipelines to enhance performance.

While both regression-based and correspondence-based approaches perform well on standard benchmarks, 
a key limitation, especially in safety-critical applications, 
is their inability to provide reliable statistical guarantees and accurate pose confidence regions. 
Inspired by~\cite{yang2023object}, we propose a deterministic 6D pose confidence region estimation approach to address these issues.

Additionally, current 6D pose evaluation metrics primarily emphasize precision assessment through geometric alignment criteria. 
The reprojection metric assesses keypoint localization accuracy by measuring compliance with predefined pixel-error thresholds,
while the 5$^{\circ}$/5cm benchmark requires simultaneous satisfaction of angular (5°) and translational (5cm) error bounds. 
For detailed geometric verification, the ADD(-S) metric computes mean keypoint displacement normalized by object diameter (10\% threshold). 
However, these conventional metrics systematically neglect the critical dimension of measurement uncertainty in pose estimation. 
To bridge this critical gap, we propose novel quantification metrics for confidence region in Section~\ref{sub_metric},
considering both of ground-truth coverage probability and volumetric compactness.

\subsection{Uncertainty Quantification of 6D Pose} 

The increasing application of DNNs in safety-critical fields has heightened the demand for reliable uncertainty quantification (UQ) methods. 
These methods primarily address two types of uncertainties: aleatoric uncertainty, which arises from data noise, and epistemic uncertainty, associated with model parameters~\cite{feng2021review}. 
Both types of uncertainties have been estimated across various visual applications, including semantic segmentation~\cite{postels2019sampling}, optical flow~\cite{ilg2018uncertainty}, keypoint detection~\cite{zhou2023star,kumar2020luvli}, and object detection~\cite{he2019bounding}. 
In this study, we focus specifically on aleatoric uncertainty.

There are three main approaches to UQ in deep learning.
Deep Ensembles train multiple networks, 
aggregating their outputs through diverse initializations and training data~\cite{shi2021fast}.
MC-Dropout, 
grounded in variational inference~\cite{Gal2016Uncertainty}, 
approximates Bayesian posterior distributions using multiple stochastic forward passes with active dropout layers. 
Both methods face scalability challenges due to high computational and memory requirements.
Another approach is Direct Modeling, 
where a probability distribution over network outputs is assumed, 
and the network's output layers directly estimate the parameters of this distribution. 

By applying Direct Modeling approaches,
recent advances in 6D pose estimation have systematically adopted the Bingham distribution for modeling rotational uncertainty.
Gilitschenski et al.~\cite{Gilitschenski2020} introduced a differentiable Bingham loss function, enabling direct learning of rotation distributions.
Building on this, Okorn et al.~\cite{okorn2020learning} demonstrated effective modeling of asymmetric object rotations through isotropic Bingham parametrization, 
while Deng et al.~\cite{deng2022deep_bingham} proposed Deep Bingham Networks to account for a family of pose hypotheses. However, these methods do not provide confidence regions for 6D pose estimation.

The most related work to ours is by Yang et al.~\cite{yang2023object}.
It uses ICP for keypoint confidence region prediction and a sampling-based method to propagate uncertainty to 6D pose. 
However, sampling methods are computationally inefficient and increase confidence region volume. 
The GPU-accelerated sampling by Gao et al.~\cite{Gao2024closure} merely re-samples the output of~\cite{yang2023object}, failing to reduce time while still producing conservative confidence regions.
So we propose a deterministic approach using the IFT to propagate keypoint confidence regions directly to pose confidence regions.
\section{Method}
As depicted in~\cref{fig_structure}, 
we present a deterministic method for 6D pose confidence region estimation. 
Starting with an input image, we regress the Gaussian distributions of keypoints (\cref{keypoint_direct_reg}) to create keypoint confidence regions (\cref{sec_keypoint_confidence_region}) that ensure a specified coverage of ground truth keypoints. 
We then apply the single-shot PnP to predict the 6D pose and use the IFT to propagate the keypoint confidence regions into 6D pose (\cref{sec_pose_confidence_region}).

\subsection{Keypoint Deterministic Regression}
\label{keypoint_direct_reg}

Heatmap-based methods, while accurate, 
require significant computational and storage resources, 
limiting their flexibility~\cite{li2021human}.
To address this, we adopt an efficient and deterministic regression-based keypoint detection approach~\cite{wang2024monocular}, 
which directly maps the input image $\mathbf{I} \in \mathbb{R}^{H\times W\times 3}$ to a set of $N$ Gaussian distributions, 
$\mathcal{P}_\Theta(\mathbf{x}_n|\mathbf{I})$, 
where $n = 1...N$ and $N$ is the total number of predefined semantic keypoints.
Each distribution $\mathcal{P}_\Theta(\mathbf{x}_n|\mathbf{I})$ represents the probability of the true keypoint $\tilde{\mathbf{x}}_n \in \mathbb{R}^2$ being at position $\mathbf{x}_n \in \mathbb{R}^2$.
To handle the aleatoric uncertainty, 
the variance term is integrated into the training loss, 
enabling the unsupervised generation of these distributions.
The objective is to optimize model parameters $\Theta$ by maximizing the likelihood of annotated keypoints. 
This approach helps account for ambiguities in labeled data, due to perspective variations and occlusion. 
We utilize the Negative Log-Likelihood (NLL) loss function to quantify the divergence between the predicted distributions and the ground-truth Dirichlet distribution~\cite{teh2017dirichlet}.
\begin{equation}
\label{equ:NLL_loss}
\mathcal{L}_{NLL} = -\sum^N_{n=1}\log \mathcal{P}_{\Theta}(\mathbf{x}_n|\mathbf{I})\bigg|_{\mathbf{x}_n=\tilde{\mathbf{x}}_n}
\end{equation}

The trained model on the train dataset $\mathcal{D}_{train}$ serves as the prediction function $F$ for conformal prediction, and is applied for keypoint confidence region estimation in the next section.
The network architecture follows the design in~\cite{wang2024monocular}, where the model outputs $F(\mathbf{I})_n$ represent a Gaussian distribution $\mathcal{N}(\mathbf{x}_n, \boldsymbol{\sigma}_n^2)$. Its probability density function (PDF) is parameterized as $\mathcal{P}_\Theta(\mathbf{x}_n|\mathbf{I})$.
Where $\mathbf{x}_n$ is the mean vector $\mathbf{x}_n$ and $\boldsymbol{\sigma}_n^2\in \mathbb{R}^{2\times 2}$ is the covariance matrix,
both are predicted through multi-layer perceptrons (MLPs).

\subsection{Conformal Keypoint Confidence Region}
\label{sec_keypoint_confidence_region}

We adopt an ICP framework to predict the conformal keypoint confidence region, 
denoted by $\mathcal{K}^{r}$ as shown in~\cref{fig_structure} (e). 
The ICP process for the trained prediction function $F$ involves two key steps: 
conformal calibration on the calibration dataset $\mathcal{D}_{cal}=\{\mathbf{I}^i,\tilde{\mathbf{x}}^{i}\}_{i=1}^l$ and conformal prediction on the test dataset $\mathcal{D}_{test}$. 
$l$ denotes the total number of samples in \( \mathcal{D}_{cal} \).
During calibration, a sequence of nonconformity scores is generated. 
These scores are applied in the prediction stage to compute $\mathcal{K}^{r}$.
Given a new sample $(\mathbf{I}^{l+1}, \tilde{\mathbf{x}}^{l+1})\in \mathcal{D}_{test}$ that satisfies the exchangeability condition~\cite{yang2023object},
ICP predicts a confidence region $\mathcal{K}^{r}(\mathbf{I}^{l+1},\epsilon)$, 
parameterized by a user-specified error rate $\epsilon$, such that:

\begin{equation}
    \label{eq:cp_probability}
    \mathbb{P}\left[\tilde{\mathbf{x}}^{l+1}\in \mathcal{K}^{r}(\mathbf{I}^{l+1},\epsilon)\right]\geq1-\epsilon,
\end{equation}
This implies that the prediction region $\mathcal{K}^{r}(\mathbf{I}^{l+1},\epsilon)$ for the new sample covers the true keypoint location $\tilde{\mathbf{x}}^{l+1}$ with a probability of at least $1-\epsilon$. 

The ICP framework addresses a bi-objective problem: 
while ensuring the coverage condition stated~\cref{eq:cp_probability}, 
the goal is to minimize the size of the prediction region $\mathcal{K}^{r}(\mathbf{I}^{l+1},\epsilon)$. 
By balancing the above two objectives, 
ICP provides a principled approach to constructing confidence regions that are both reliable and as tight as possible.

The most critical aspect of ICP conformal calibration is designing the nonconformity function, 
which is denoted by $r$.
It is constructed to quantify the discrepancy between the new sample and the training dataset $\mathcal{D}_{train}$.
The neural network trained using $\mathcal{D}_{train}$ is referred to as the function  $F$.
Hence, the  nonconformity function can be denoted as:
\begin{equation} \label{eq:non_conformal_score}
    r(\boldsymbol{\tilde{\mathbf{x}}}, F(\mathbf{I}))=\max 
    \left\{
    \phi(\tilde{\mathbf{x}}_n, F(\mathbf{I})_n)
    \right\}_{n=1}^N
\end{equation}
where the function $\phi$ computes the nonconformity score for each keypoint, 
with the highest score representing the overall score.
For the $n$-th keypoint, 
$\phi$ is defined as follows:
\begin{equation}
    \label{eq:phi_before}
    \phi(\tilde{\mathbf{x}}_n, F(\mathbf{I})_n)=
    \frac{\|\tilde{\mathbf{x}}_n - \mathbf{x}_n\|}{\mathrm{det}(\boldsymbol{\sigma}_n)}
\end{equation}

For the $i$-th sample in the calibration set $\mathcal D_{cal}$, 
the nonconformity score, calculated using~\cref{eq:non_conformal_score}, 
is denoted as \( \alpha_{i} = r(\boldsymbol{\tilde{\mathbf{x}}}^{i}, F(\mathbf{I}^{i})) \) for \( i = 1, \dots, l \). 
These scores are then ranked in descending order as $\alpha_{\psi(1)} \geq \dots \geq \alpha_{\psi(l)}$, 
where $\psi(\cdot)$ denotes the ranking index. 
Next, given $\epsilon \in (0,1)$, 
the $\lfloor l\epsilon\rfloor$-th largest nonconformity score is denoted as $\alpha_{\psi(\lfloor l\epsilon\rfloor)}$. 
Using this score and considering~\cref{eq:cp_probability} and~\cref{eq:non_conformal_score}, 
the keypoint confidence regions $\mathcal{K}^{r}(\mathbf{I}^{l+1}, \epsilon)$ of the newly introduced test image \(\mathbf{I}^{l+1}\) from \(\mathcal{D}_{\text{test}}\) is derived as:
\begin{equation}
\label{eq:ICP_set}
\begin{split}
&\mathcal{K}^{r}(\mathbf{I}^{l+1}, \epsilon)\\
&=\{\mathbf{x}'\mid\max\{\phi(\mathbf{x}_n',F(\mathbf{I}^{l+1})_n)\}_{n=1}^{N}\leq\alpha_{\psi(\lfloor l\epsilon\rfloor)}\}
\\ &=\{\mathbf{x}'\mid\phi(\mathbf{x}_n',F(\mathbf{I}^{l+1})_n)\leq\alpha_{\psi(\lfloor l\epsilon\rfloor)},\forall n\},
  \end{split}
\end{equation}
Note that $\max\{\phi_1,\ldots,\phi_n,\ldots,\phi_N\}\leq\alpha$ holds if and only if $\phi_n\leq\alpha$ for every $n$.
We then substitute~\cref{eq:phi_before} into~\cref{eq:ICP_set} and arrive at the desired result.
\begin{equation}
\label{eq_ICP_set_ball}
\mathcal{K}^{r}(\mathbf{I}^{l+1}, \epsilon)=\left\{\mathbf{x}'\mid \|\mathbf{x}_n' - \mathbf{x}_n\| \leq \mathrm{det}(\mathbf{\sigma}_n)\alpha_{\psi(\lfloor l\epsilon\rfloor)},\forall n\right\}
\end{equation}

~\cref{eq_ICP_set_ball} describes an ball region for the $k$-th keypoint centered at $\mathbf{x}_n$. 
The area of the ball is proportional to the determinant of the covariance matrix $\mathrm{det}(\sigma_{n})$ and a scaling factor $\alpha_{\psi(\lfloor l\epsilon\rfloor)}$. 
Note that the size of the confidence region increases when the keypoint's Gaussian distributions are uncertain.
It can be indicated by a large determinant in the covariance matrix, 
and the kepoint regression perform poorly on $\mathcal{D}_{cal}$, 
leading to a large $\alpha_{\psi(\lfloor l\epsilon\rfloor)}$.

\begin{figure*}[ht]
    \includegraphics[width=\textwidth]{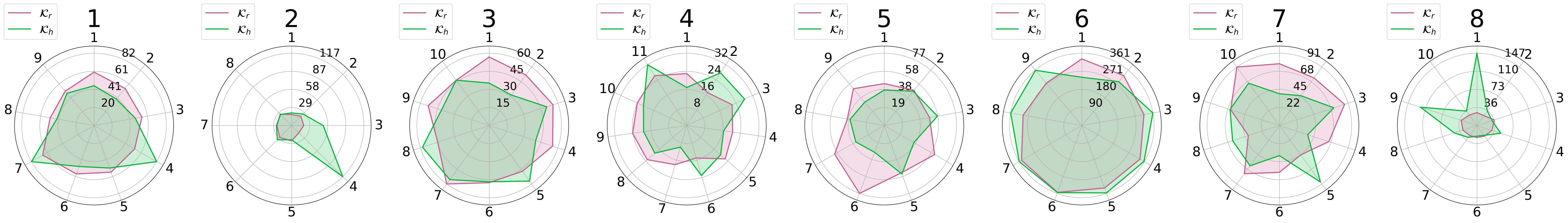}  
    \caption{Radar charts illustrating the mean radii of keypoint confidence regions ($\mathcal{K}^{h}$ and $\mathcal{K}^{r}$). 
    The eight radar charts correspond to the eight objects in the LMO dataset (from left to right). 
    Each axis in a given radar chart represents one of the specific object's keypoints.}
    \label{fig:rose_fig}
\end{figure*}

The distinction between pose confidence region $\mathcal{K}^{r}$ proposed in this paper and those by Yang et al. $\mathcal{K}^{h}$~\cite{yang2023object} lies in our introduction of a novel nonconformity function \(\phi\) related to $\mathcal{P}_\Theta(\mathbf{x}_n|\mathbf{I})$ in~\cref{keypoint_direct_reg}.
As shown in~\cref{tab_time}, $\mathcal{K}^{r}$ is generated faster compared to $\mathcal{K}^{h}$. 
Additionally, $\mathcal{K}^{r}$ is more stable, 
avoiding the generation of excessively large confidence regions compared with $\mathcal{K}^{h}$ (e.g., keypoint 4 of object 2 and keypoints 1 and 9 of object 8 in~\cref{fig:rose_fig}).

\subsection{Deterministic Pose Confidence Region}
\label{sec_pose_confidence_region}

For propagating the $\mathcal{K}^{h}$ to pose confidence region $\mathcal{C}^{s}_{h}$,
Yang et. al.~\cite{yang2023object} propose an inefficient sampling algorithm, 
named Random Sampling Averaging. 
They first sample three keypoints from $\mathcal{K}^{h}$ and then solve the P3P problem to obtain the pose sample. 
Next, the convex hull of pose samples whose reprojected keypoints all fall within $\mathcal{K}^{h}$ are considered as $\mathcal{C}^{s}_{h}$.
However, the sampling-based approach presents several challenges. 
For instance, as the number of trials increases, the sampling time grows significantly. 
Additionally, in challenging scenarios, the number of valid samples becomes insufficient. 
Moreover, the pose estimation accuracy of the P3P algorithm tends to be lower, causing the pose samples to deviate further from the ground truth, which in turn enlarges the volume of $\mathcal{C}^{s}_{h}$.

To address the above issues, as shown in~\cref{fig_structure} (f), 
we calculate the Jacobians of 6D pose with respect to the 2D keypoints based on implicit function theorem (IFT)~\cite{krantz2002implicit}.
Then we deterministicly propagate $\mathcal{K}^{r}$ to $\mathcal{C}^{d}_{r}$ using the uncertainty propagation theorem.
The Jacobian is difficult to compute due to the non-linear relationship in perspective geometry. 
The IFT implicitly computes Jacobians by leveraging geometric constraints, 
avoiding the need for an explicit solution.
Following~\cite{chen2020end}, we define a single-shot PnP solver, which outputs the pose $\mathbf{y}=[y_1,...,y_m]^T$, denoted as $g$:
\begin{equation}
  \mathbf{y}=g(\mathbf{x},\mathbf{z},\textbf{K})
  \label{eq:uncertain_propagation}
\end{equation}
where \( \mathbf{x}\in \mathbb{R}^{2\times N}\) and \( \mathbf{z}\in \mathbb{R}^{3\times N} \) represent $N$ 2D-3D correspondences, 
and \( \textbf{K}\in \mathbb{R}^{3\times 3} \) is the camera intrinsic matrix.
The objective function of $g$ is defined as follow:
\begin{equation}
    O(\mathbf{x},\mathbf{y},\mathbf{z},\mathbf{K})=\sum^N_{n=1}\|\mathbf{r}_n\|^2_2
  \label{eq:uncertain_propagation}
\end{equation}
where $\mathbf{r}_n=\mathbf{x}_n-\boldsymbol{\pi}_n$ is the reprojection error of the $n$-th correspondence. 
\( \boldsymbol{\pi}_n = \Pi(\mathbf{z}_n|\mathbf{y},\mathbf{K}) \) represents the reprojected points, calculated by the projective function \(\Pi\).

A stationary condition can be formulated by computing the first-order derivative of $O$ with respect to \( \mathbf{y} \).

\begin{equation}
  \frac{\partial O(\mathbf{x},\mathbf{y},\mathbf{z},\mathbf{K})}{\partial \mathbf{y}}=0
  \label{eq:constraint_function}
\end{equation}
The IFT constraint function $f$ is constructed based on~\cref{eq:constraint_function}.
\begin{equation}
    f(\mathbf{x},\mathbf{y},\mathbf{z},\mathbf{K})=[f_1,...,f_m]^T
  \label{eq:constraint_function2}
\end{equation}
The dimensionality of the pose, i.e., $m$, depends on the parameterization of the rotation $SO(3)$. 
Chen et al.~\cite{chen2020end} use the less intuitive axis-angle representation $m=6$. 
For better user comprehension, we adopt the Euler angle in this work and  $m=6$.  
For each parameter of the pose representation, the constraint function $f_j$ is expressed as:
\begin{equation}
    f_j=\frac{\partial O (\mathbf{x},\mathbf{y},\mathbf{z},\mathbf{K})}{\partial y_j} = 2\sum_{n=1}^N\langle \mathbf{r}_n,-2\frac{\partial \boldsymbol{\pi}_n}{\partial y_j}\rangle 
  \label{eq:constraint_function3}
\end{equation}
Furthermore, based on the IFT, we can apply the constraint function \( f \) to compute \( \frac{\partial g}{\partial{\mathbf{x}}} \).

\begin{equation}
    \frac{\partial g}{\partial\mathbf{x}}=-\left[\frac{\partial f}{\partial\mathbf{y}}\right]^{-1}\left[\frac{\partial f}{\partial\mathbf{x}}\right]
\end{equation}

Subsequently, the covariance matrix \(\mathbf{\Sigma_\mathbf{x}} \in \mathbb{R}^{2N \times 2N}\) representing $\mathcal{K}^{r}$ is directly propagated to \(\mathbf{\Sigma_\mathbf{y}} \in \mathbb{R}^{6 \times 6}\) standing for $\mathcal{C}^{d}_{r}$ using the uncertainty propagation theorem.
\begin{equation}
  \mathbf{\Sigma}_{\mathbf{y}}=\frac{\partial g}{\partial{\mathbf{x}}}\mathbf{\Sigma}_{\mathbf{x}}\frac{\partial g}{\partial{\mathbf{x}}}^T
  \label{eq:uncertain_propagation}
\end{equation}
The covariance matrices for the Euler angles and the translation vector are denoted as $\mathbf{\Sigma}_{\mathbf{R}}=\mathbf{\Sigma}_{\mathbf{y}}(1:3,1:3)$ and $\mathbf{\Sigma}_{\mathbf{t}}=\mathbf{\Sigma}_{\mathbf{y}}(4:6,4:6)$, respectively.

In summary, as shown in~\cref{fig_structure}, our method eliminates the need for inefficient sampling and integration processes by developing a deterministic method for tighter 6D pose confidence regions \(\mathcal{C}^{d}_{r}=\{\mathcal{R}^{d}_{r},\mathcal{T}^{d}_{r}\}\).
\begin{align}
\mathcal{R}^{d}_{r} &= \{\mathbf{x} \in \mathbb{R}^3 :
( \mathbf{x}-\mathbf{y}_{1:3})^\top \mathbf{\Sigma}_\mathbf{R}^{-1}
( \mathbf{x}-\mathbf{y}_{1:3}) \leq 1\} \\
\mathcal{T}^{d}_{r} &= \{\mathbf{x} \in \mathbb{R}^3 : 
( \mathbf{x}-\mathbf{y}_{4:6})^\top \mathbf{\Sigma}^{-1}_\mathbf{t}
( \mathbf{x}-\mathbf{y}_{4:6})\leq 1\}
\end{align}
Unlike the pose sample convex hull $\mathcal{C}^{s}_{h}$ obtained through sampling, 
we derived a parameterized pose confidence region based on the IFT. 
The confidence regions of rotation and translation can be represented as ellipsoids.

\section{Experiment}

\subsection{Datasets and implementation details}
We conduct experiments on the LMO~\cite{lmo_dataset_BrachmannKMGSR14} and the SPEED~\cite{Kisantal2019Satellite} datasets.  
Both are designed for 6D pose estimation task. 
LMO includes photorealistic rendered training images of randomly cluttered scenes. 
We split the dataset following~\cite{yang2023object}, 
with 200 images allocated to $\mathcal{D}_{cal}$ and 1,214 images to $\mathcal{D}_{test}$.
LMO includes 8 labeled objects with significant occlusion.
SPEED is used to test the safety-critical applicability of the proposed pose confidence region. 
To ensure a rigorous analysis, we apply a sixfold cross-validation (CV) method, dividing the 12,000 simulated images into six subsets. 
Five-sixths are designated as $\mathcal{D}_{train}$, while one-sixth is split equally into $\mathcal{D}_{cal}$ and $\mathcal{D}_{test}$. 

The runtimes are evaluated on a workstation equipped with Nvidia A6000 GPUs. 
The network architecture is based on~\cite{wang2024monocular}. 
The regression model undergoes 96 epochs of training, 
utilizing a pre-trained model based on the COCO dataset~\cite{lin2014microsoft}.
The training employs the AdamW optimizer~\cite{loshchilov2019decoupled}. 
The fixed learning rate is set as $3\times10^{-5}$.
The upper bound for the keypoint confidence region radius is set to the image diagonal length, 
as this value sufficiently covers the entire image, making larger radii unnecessary.

\subsection{Evaluation metrics}
\label{sub_metric}
We adopt keypoint reprojection accuracy~\cite{peng2019pvnet}, 
denoted by $\text{Acc}$, to evaluate the accuracy.
However, the existing metrics for pose estimation mainly focus on accuracy, neglecting the evaluation of confidence regions. 
Hence, we propose 4 metrics beyond accuracy, 
including the size of the confidence regions and coverage rate. 

\textbf{2D keypoint coverage rate:}
The validity of the region is assessed by the probability of all ground truth keypoints $\tilde{\mathbf{x}}$ falling within $\mathcal{K}^{h}$ or  $\mathcal{K}^{r}$ aiming to align with $1 - \epsilon$, as stated in~\eqref{eq:cp_probability}. 
For example, the coverage rate $\eta^{kpt}$ of $\mathcal{K}^{h}$ for $\mathcal{D}_{test}$ containing $K$ images is given as follows:
\begin{equation}
    \label{eq:kpt_cover}
    \eta^{kpt} = \frac{1}{K} \sum_{k=1}^{K} 
    \mathbb{I} \left( 
    \tilde{\mathbf{x}}^{k}\in \mathcal{K}^{h}(\mathbf{I}^{k},\epsilon) \right)
\end{equation}

\textbf{2D keypoint confidence region radius:}
We denote the region size by the keypoint confidence region radius, \(  \det(\boldsymbol{\sigma}_n)\alpha_{\psi(\lfloor l\epsilon\rfloor)}\) in~\cref{eq_ICP_set_ball}. 
The greater keypoints' uncertainty, the larger the confidence region radius.

\textbf{6D pose coverage rate:} 
Yang et al.~\cite{yang2023object} employ 2D points sampled within \(\mathcal{K}^{h}\) and use the PnP method to obtain pose samples, 
the convex hull of these samples formed a 6D pose confidence region $\mathcal{C}^{s}_{h}=\{\mathcal{R}^{s}_{h},\mathcal{T}^{s}_{h}\}$,
as shown in~\cref{fig:structure_top}. 
In contrast, we deterministicly propagate \(\mathcal{K}^{r}\) to ellipsoidal pose confidence region $\mathcal{C}^{d}_{r}=\{\mathcal{R}^{d}_{r},\mathcal{T}^{d}_{r}\}$ based on IFT.
For both Euler angles and translation vectors, 
we calculate the coverage rate of the ground truth.
Taking \(\mathcal{C}^{s}_{h}\) as an example, the coverage rate for the Euler angles $\eta^{\mathbf{R}}$ is as follows:
\begin{equation}
    \label{eq:pose_cover}
    \eta^{\mathbf{R}} = \frac{1}{K} \sum_{k=1}^{K} 
    \mathbb{I} \left( 
    \mathbf{R}\in \mathcal{R}^{s}_h(\mathbf{I}^{k},\epsilon) \right)
\end{equation}
For translation coverage rate $\eta^{\mathbf{t}}$ follows a similar form.

\begin{figure*}[htbp]
    \centering
    
    \begin{subfigure}{0.16\textwidth}
        \centering
        \includegraphics[width=\textwidth]{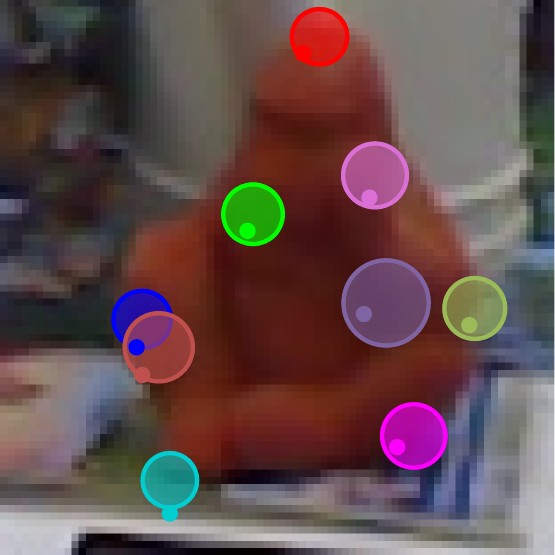}
        \caption{$\mathcal{K}^r$}
    \end{subfigure}
    \hfill
    \begin{subfigure}{0.19\textwidth}
        \centering
        \includegraphics[width=\textwidth]{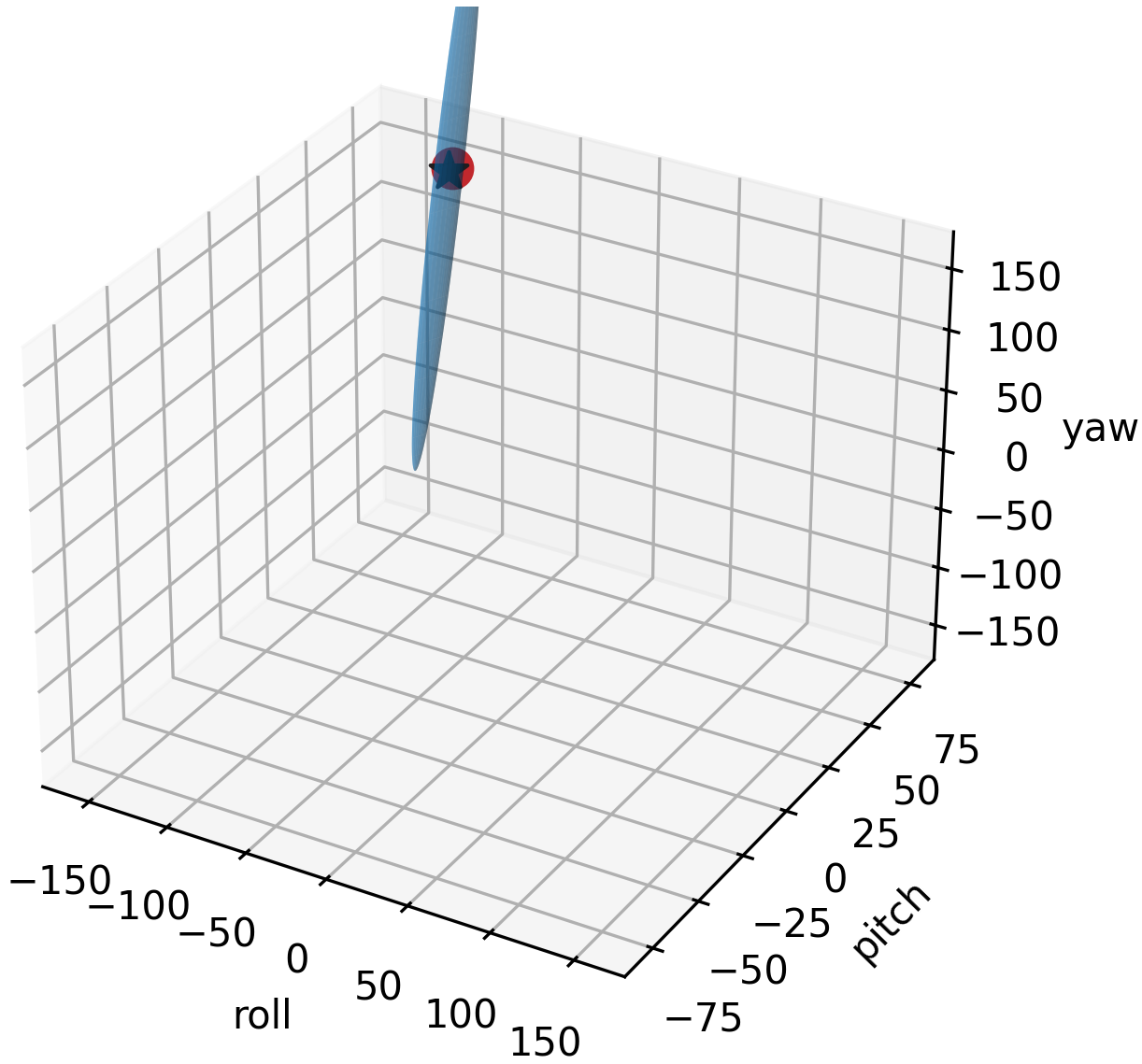}
        \caption{$\mathcal{R}^d_r:V_{\mathbf{R}}=19.6$}

    \end{subfigure}
    \hfill
    \begin{subfigure}{0.19\textwidth}
        \centering
        \includegraphics[width=\textwidth]{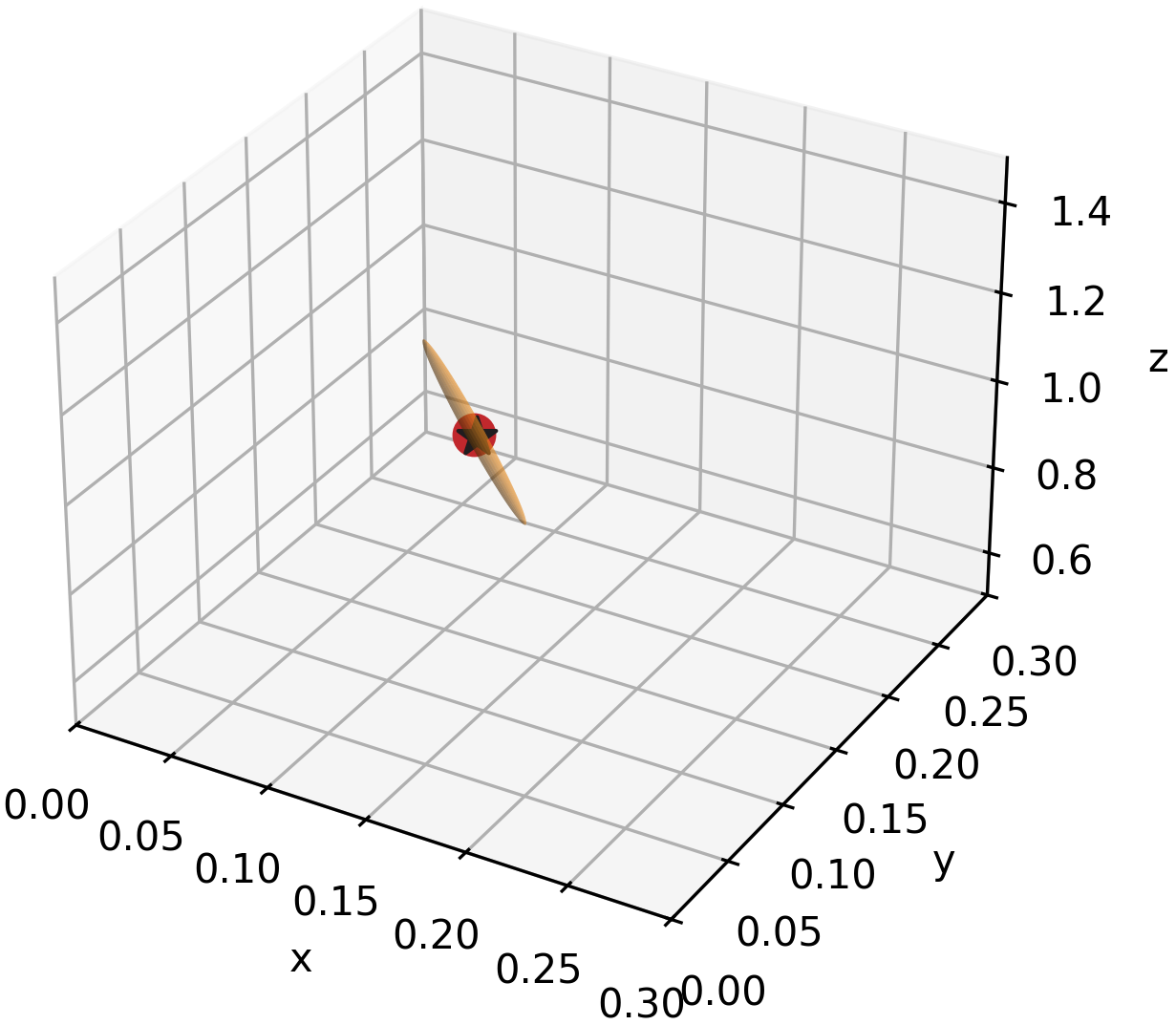}
        \caption{$\mathcal{T}^d_r:V_{\mathbf{t}}=0.05$}
    \end{subfigure}
    \hfill
    \begin{subfigure}{0.19\textwidth}
        \centering
        \includegraphics[width=\textwidth]{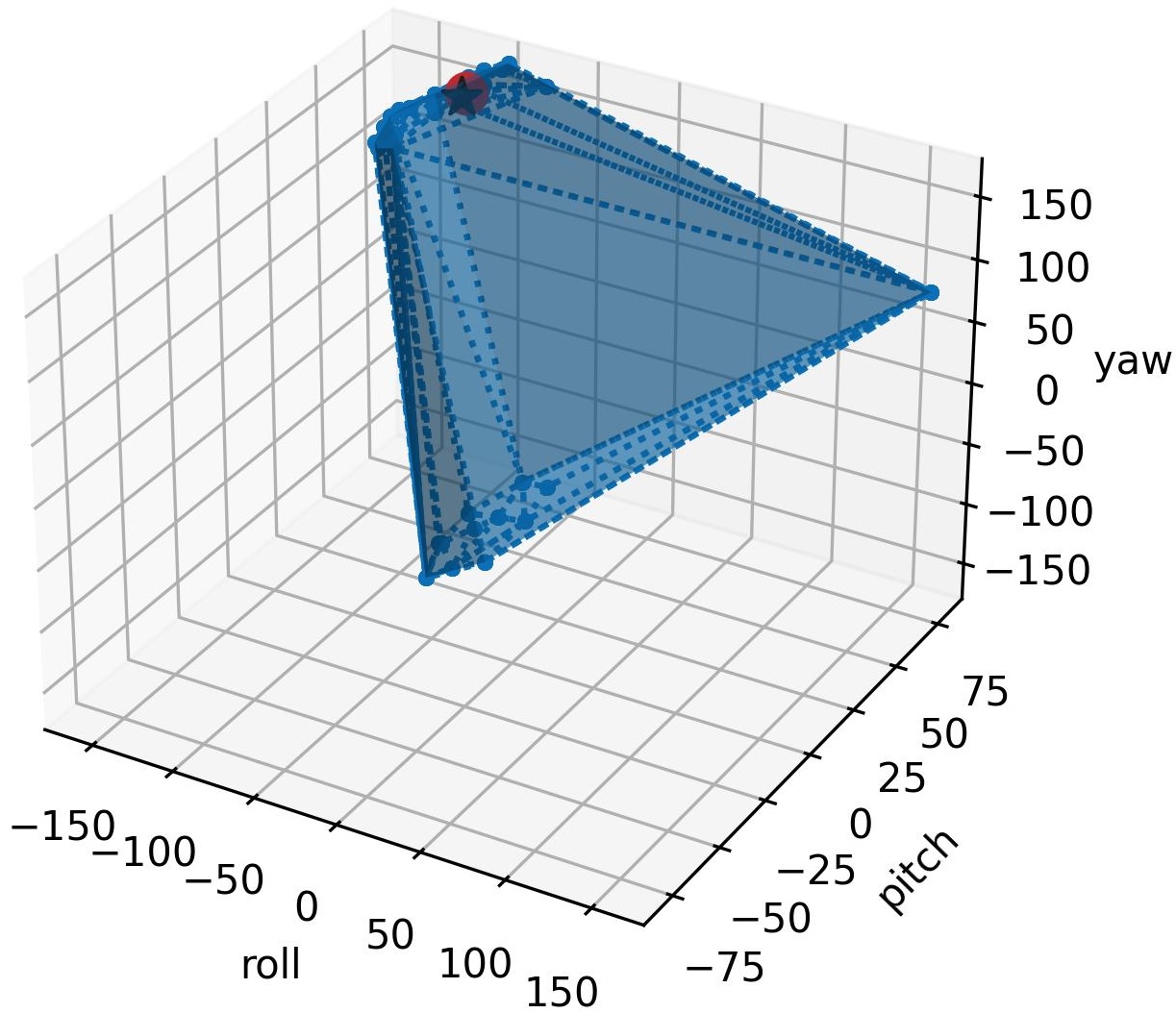}
        \caption{$\mathcal{R}^s_h:V_{\mathbf{R}}=2479.6$}
    \end{subfigure}
    \hfill 
    \begin{subfigure}{0.19\textwidth}
        \centering
        \includegraphics[width=\textwidth]{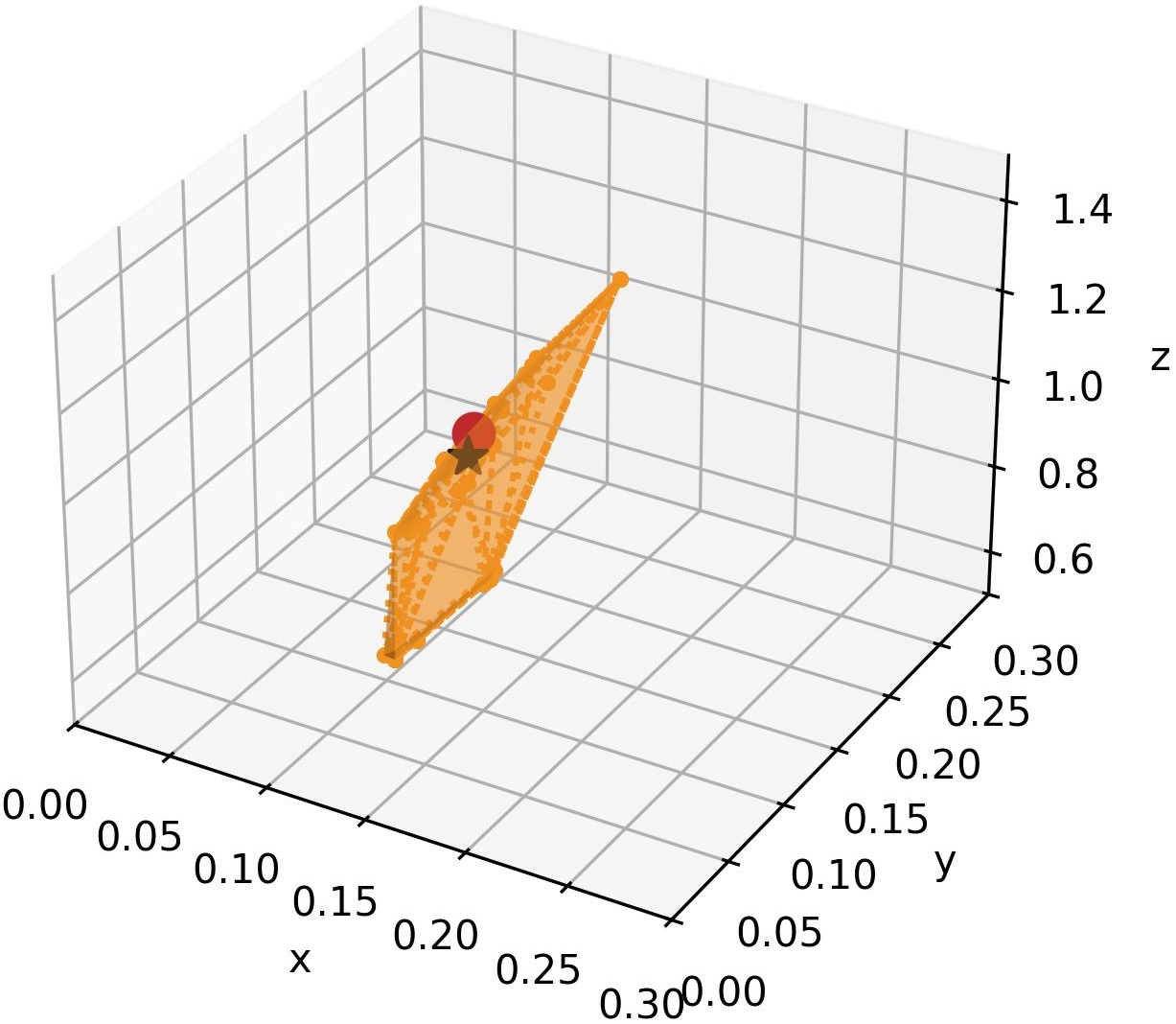}
        \caption{$\mathcal{T}^s_h:V_{\mathbf{t}}=0.5$}
    \end{subfigure}
    \hfill
    \begin{subfigure}{0.16\textwidth}
        \centering
        \includegraphics[width=\textwidth]{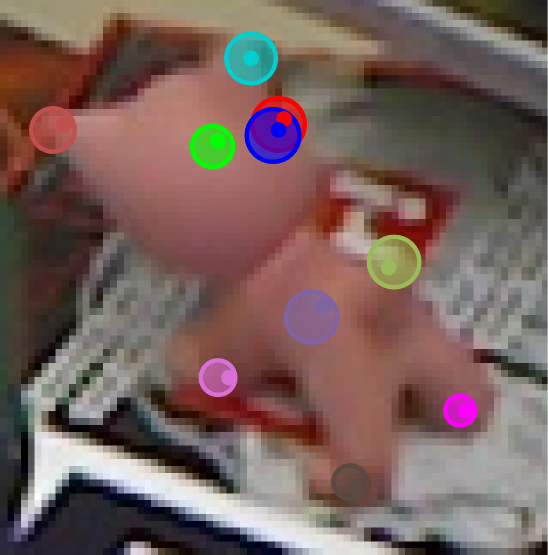}
        \caption{$\mathcal{K}^r$}
    \end{subfigure}
    \hfill
    \begin{subfigure}{0.19\textwidth}
        \centering
        \includegraphics[width=\textwidth]{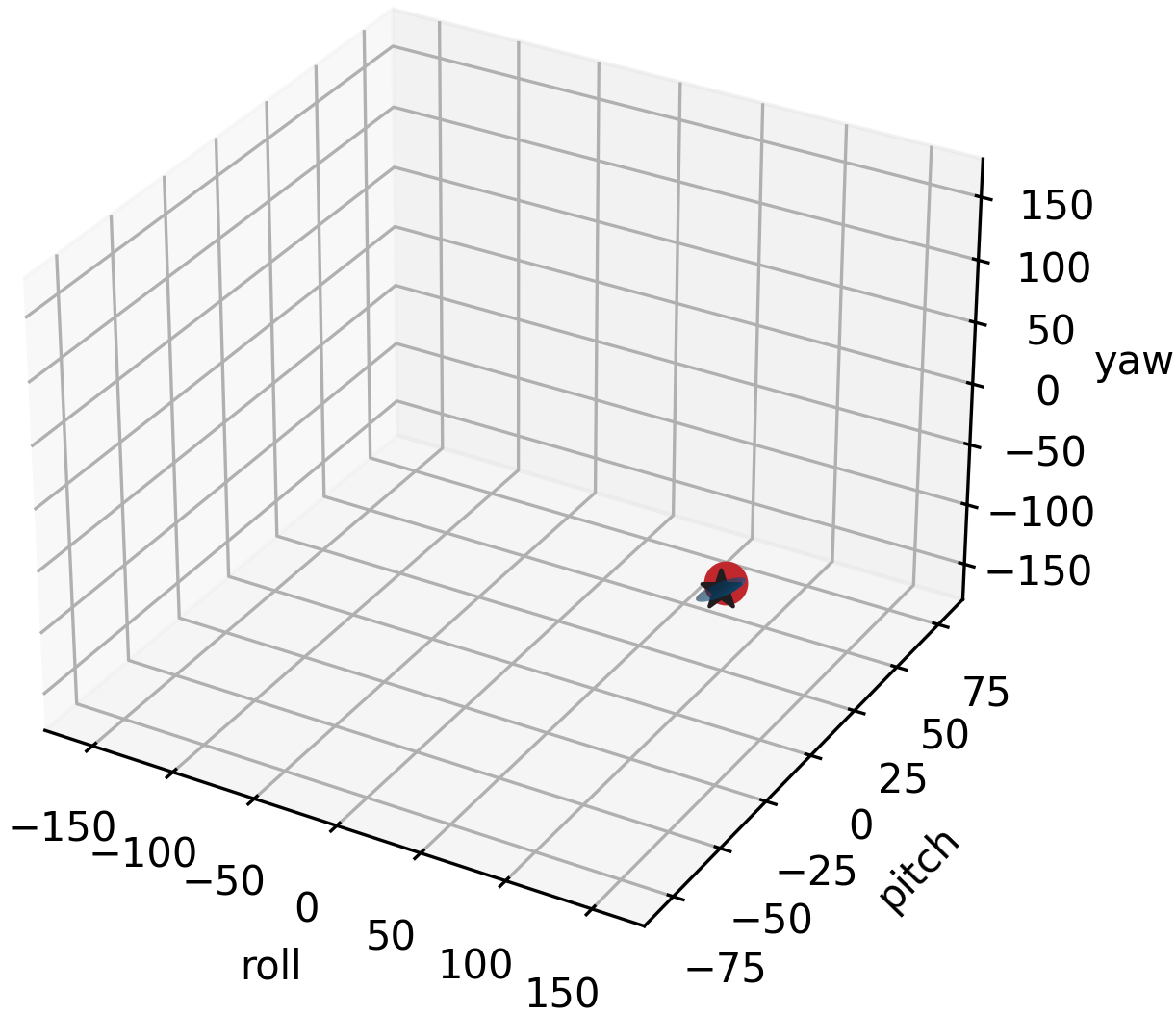}
        \caption{$\mathcal{R}^d_r:V_{\mathbf{R}}=1.0$}
    \end{subfigure}
    \hfill
    \begin{subfigure}{0.19\textwidth}
        \centering
        \includegraphics[width=\textwidth]{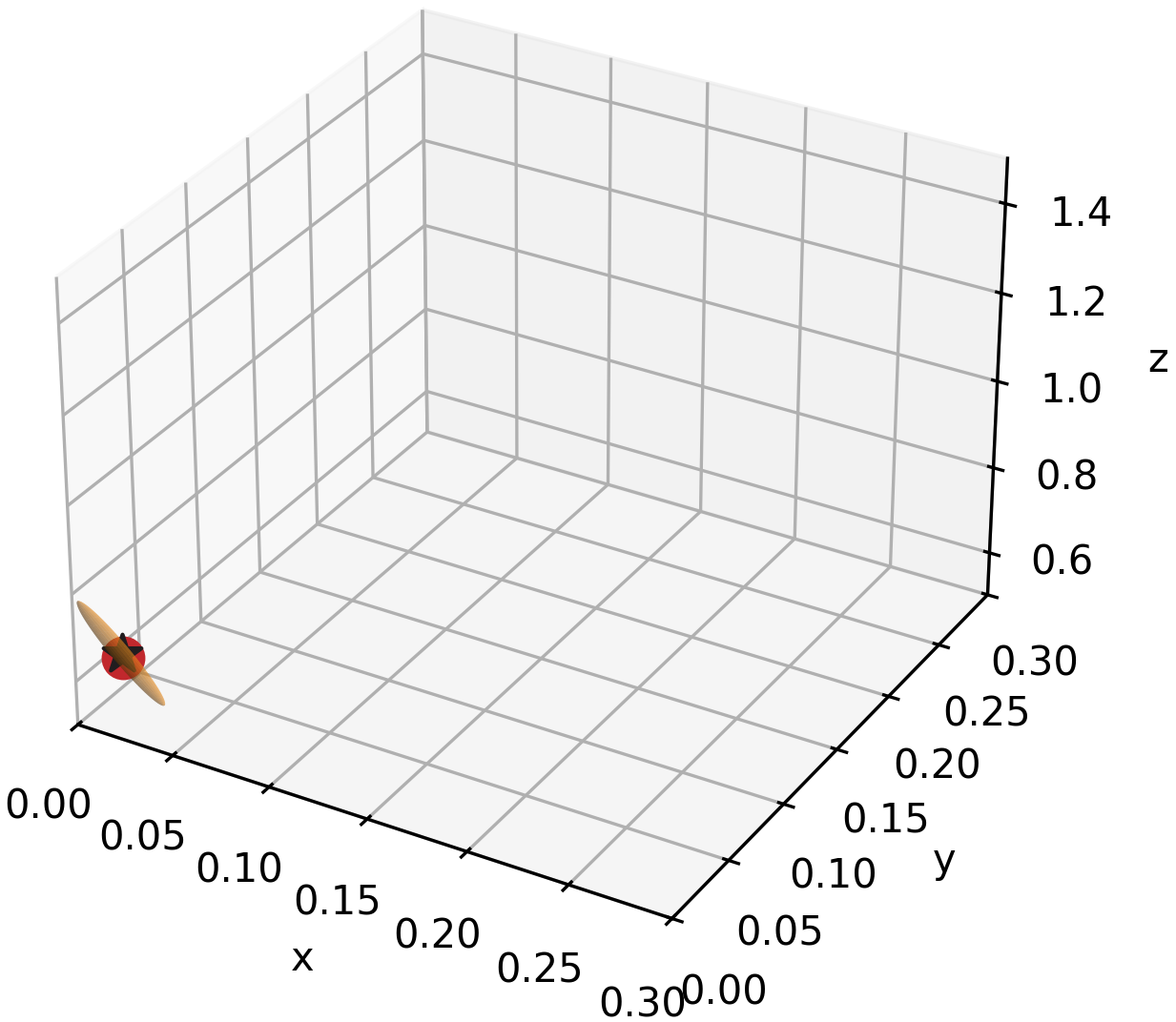}
        \caption{$\mathcal{T}^d_r:V_{\mathbf{t}}=0.01$}
    \end{subfigure}
    \hfill
    \begin{subfigure}{0.19\textwidth}
        \centering
        \includegraphics[width=\textwidth]{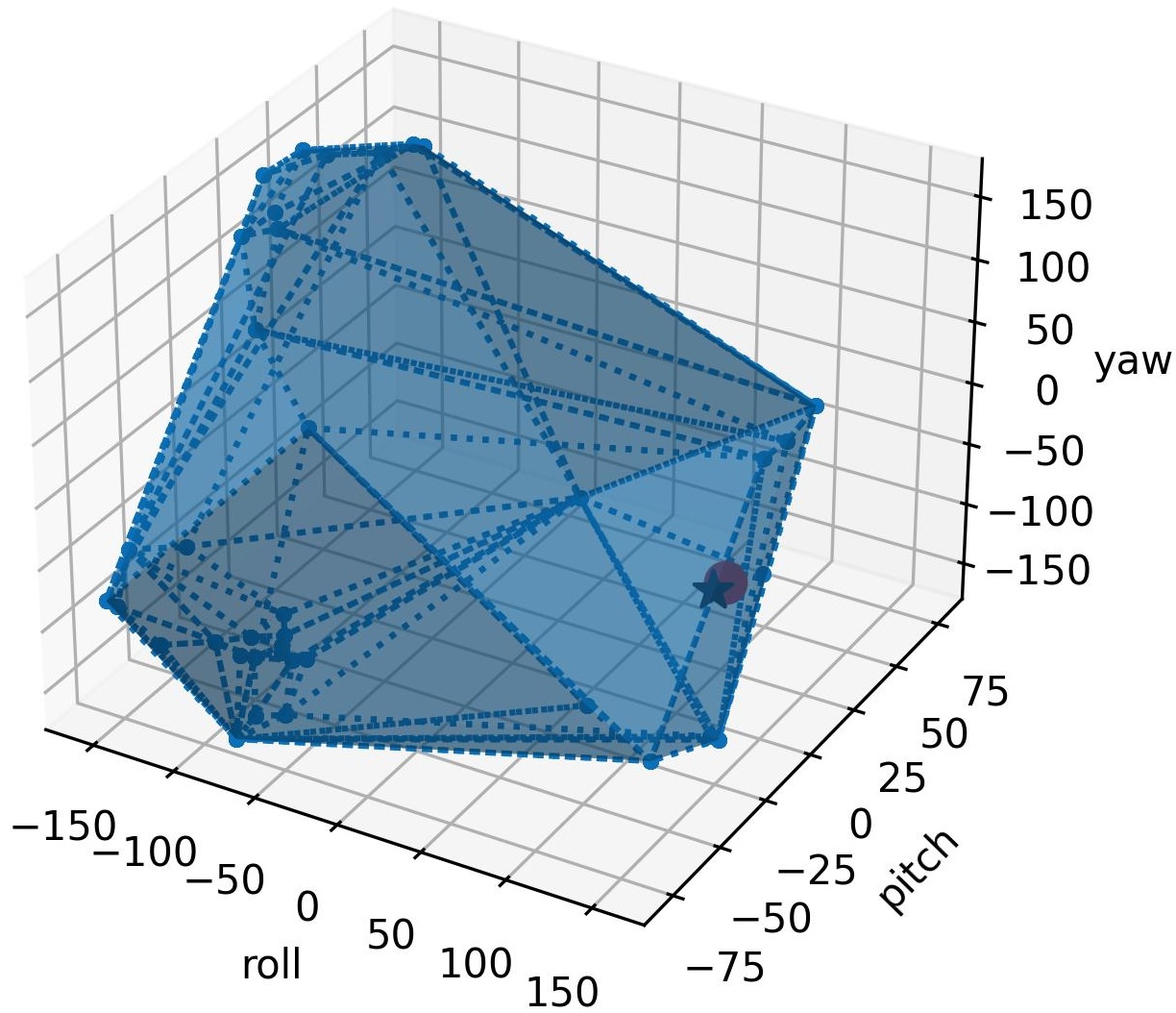}
        \caption{$\mathcal{R}^s_h:V_{\mathbf{R}}=8472.9$}
    \end{subfigure}
    \hfill
    \begin{subfigure}{0.19\textwidth}
        \centering
        \includegraphics[width=\textwidth]{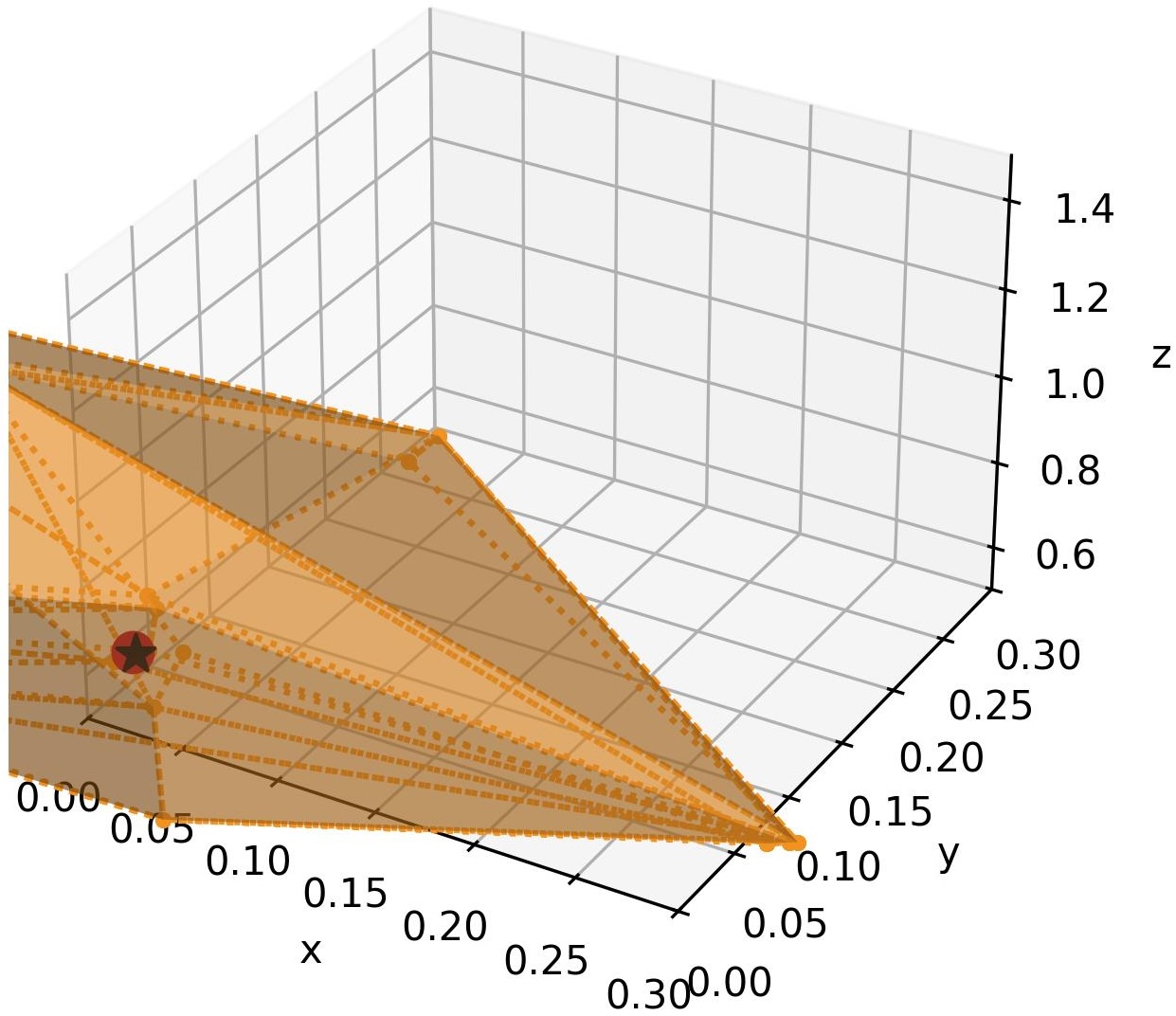}
        \caption{$\mathcal{T}^s_h:V_{\mathbf{t}}=53.8$}
    \end{subfigure}
    \hfill
    \begin{subfigure}{0.16\textwidth}
        \centering
        \includegraphics[width=\textwidth]{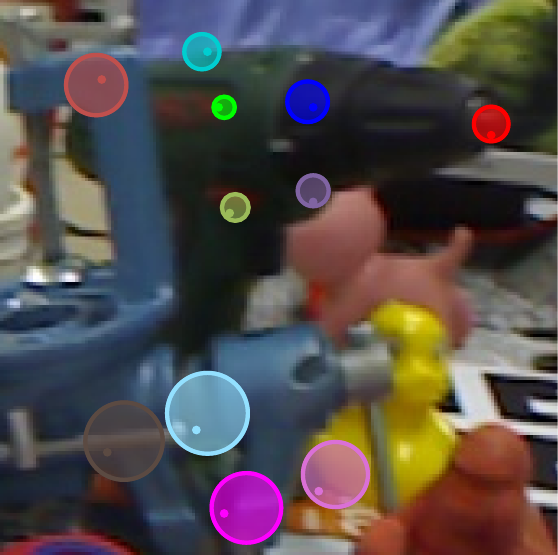}
        \caption{$\mathcal{K}^r$}
    \end{subfigure}
    \hfill
    \begin{subfigure}{0.19\textwidth}
        \centering
        \includegraphics[width=\textwidth]{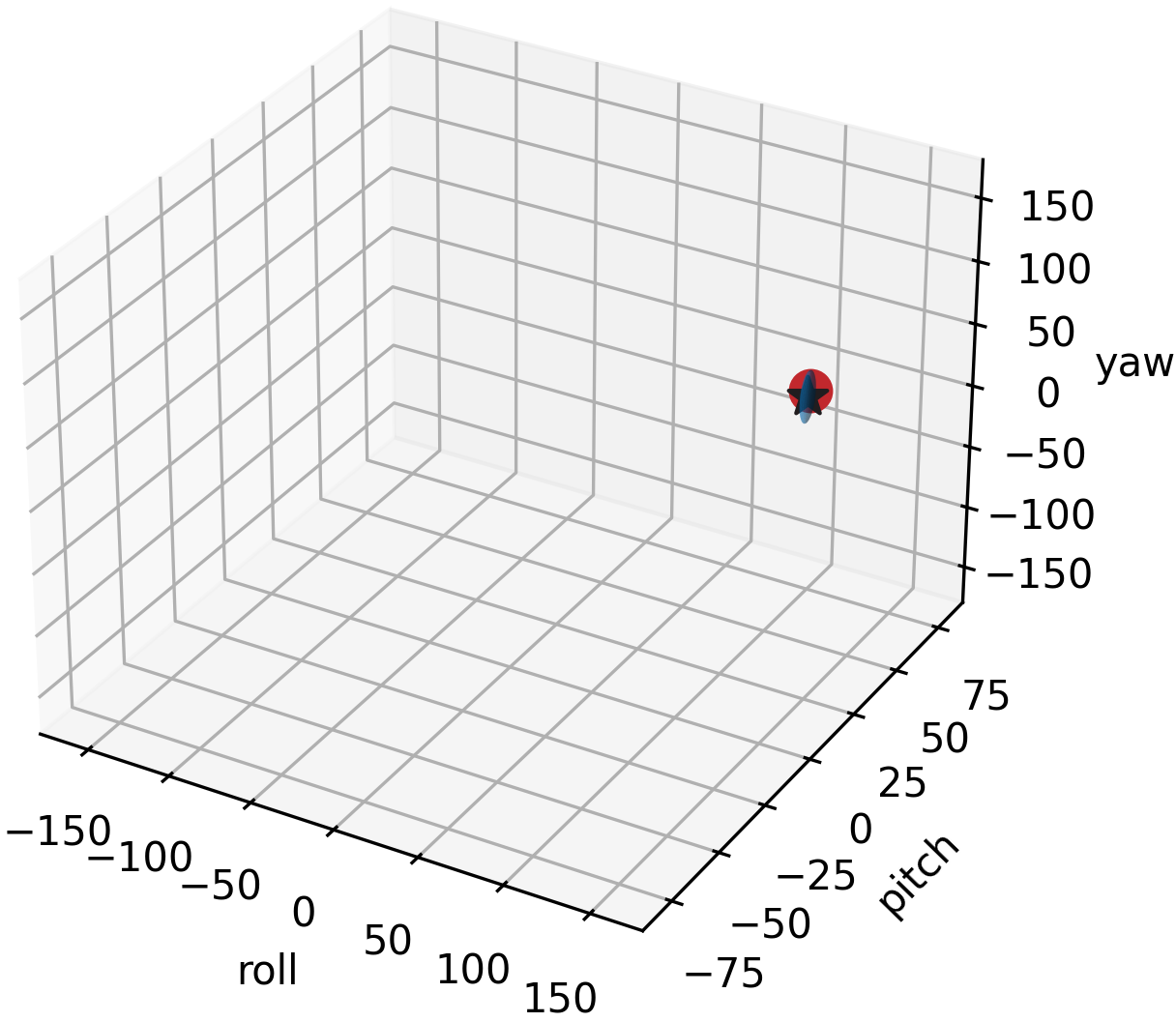}
        \caption{$\mathcal{R}^d_r:V_{\mathbf{R}}=1.9$}
    \end{subfigure}
    \hfill
    \begin{subfigure}{0.19\textwidth}
        \centering
        \includegraphics[width=\textwidth]{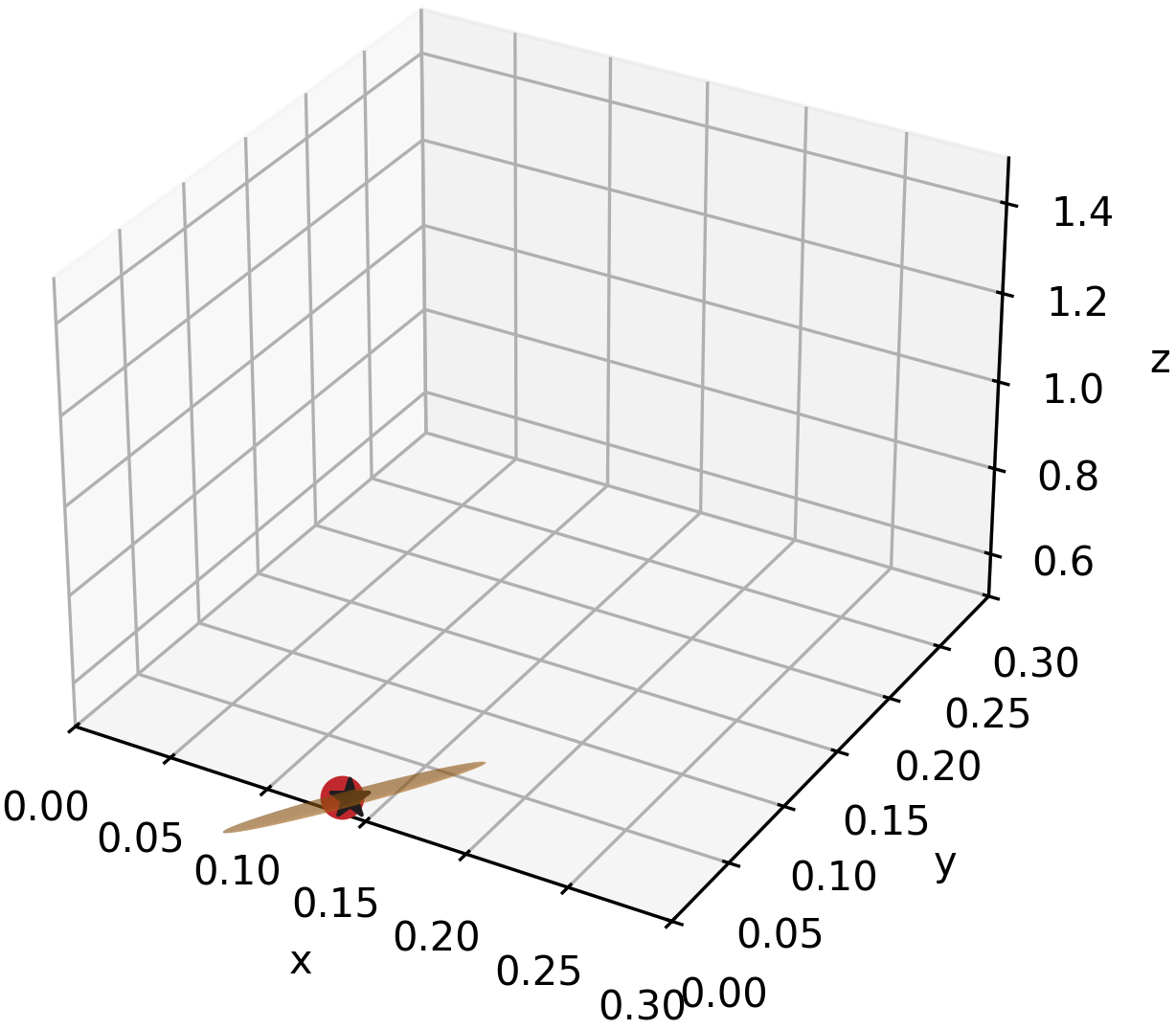}
        \caption{$\mathcal{T}^d_r:V_{\mathbf{t}}=0.03$}
    \end{subfigure}
    \hfill
    \begin{subfigure}{0.19\textwidth}
        \centering
        \includegraphics[width=\textwidth]{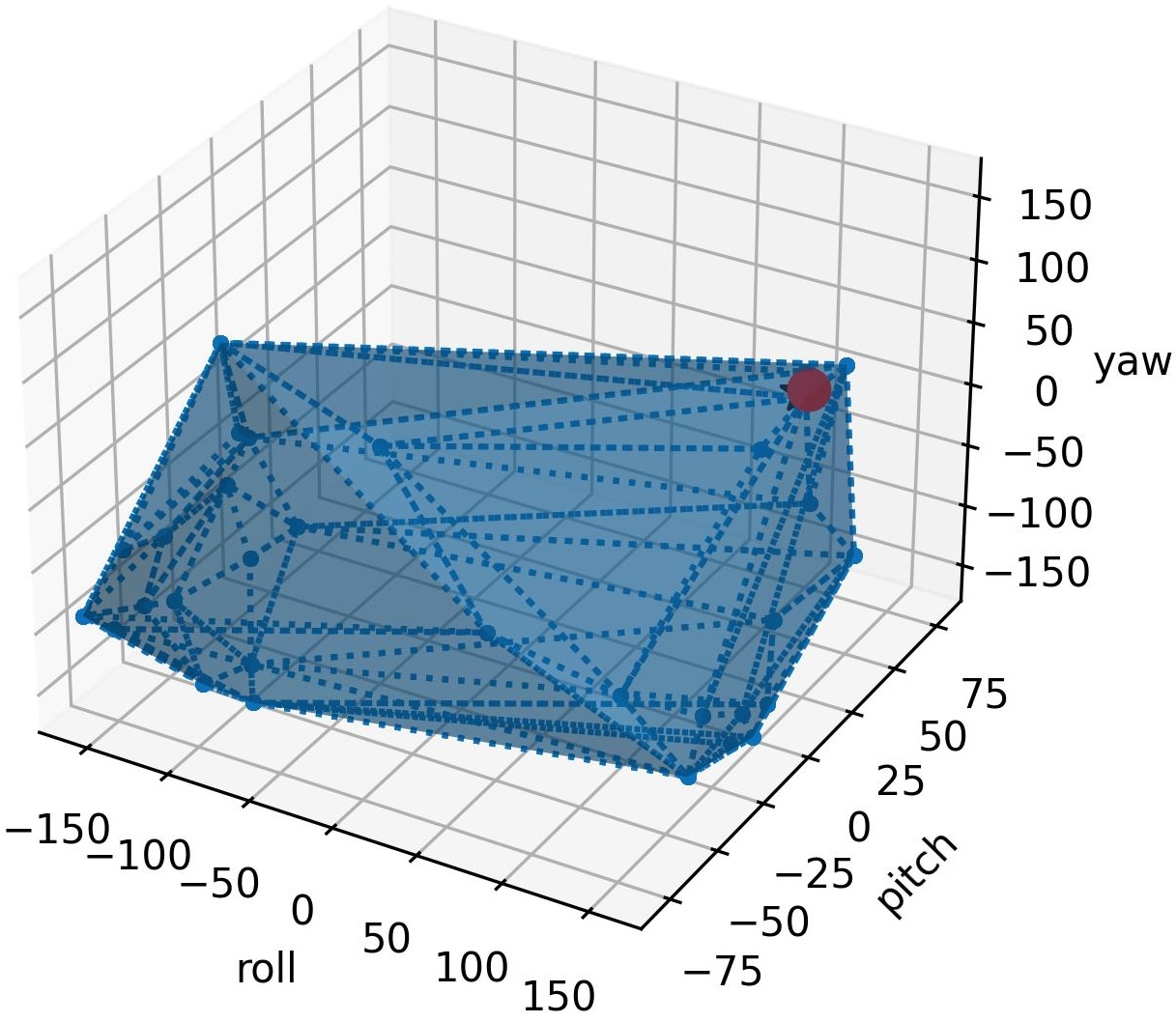}
        \caption{$\mathcal{R}^s_h:V_{\mathbf{R}}=5779.0$}
    \end{subfigure}
    \hfill
    \begin{subfigure}{0.19\textwidth}
        \centering
        \includegraphics[width=\textwidth]{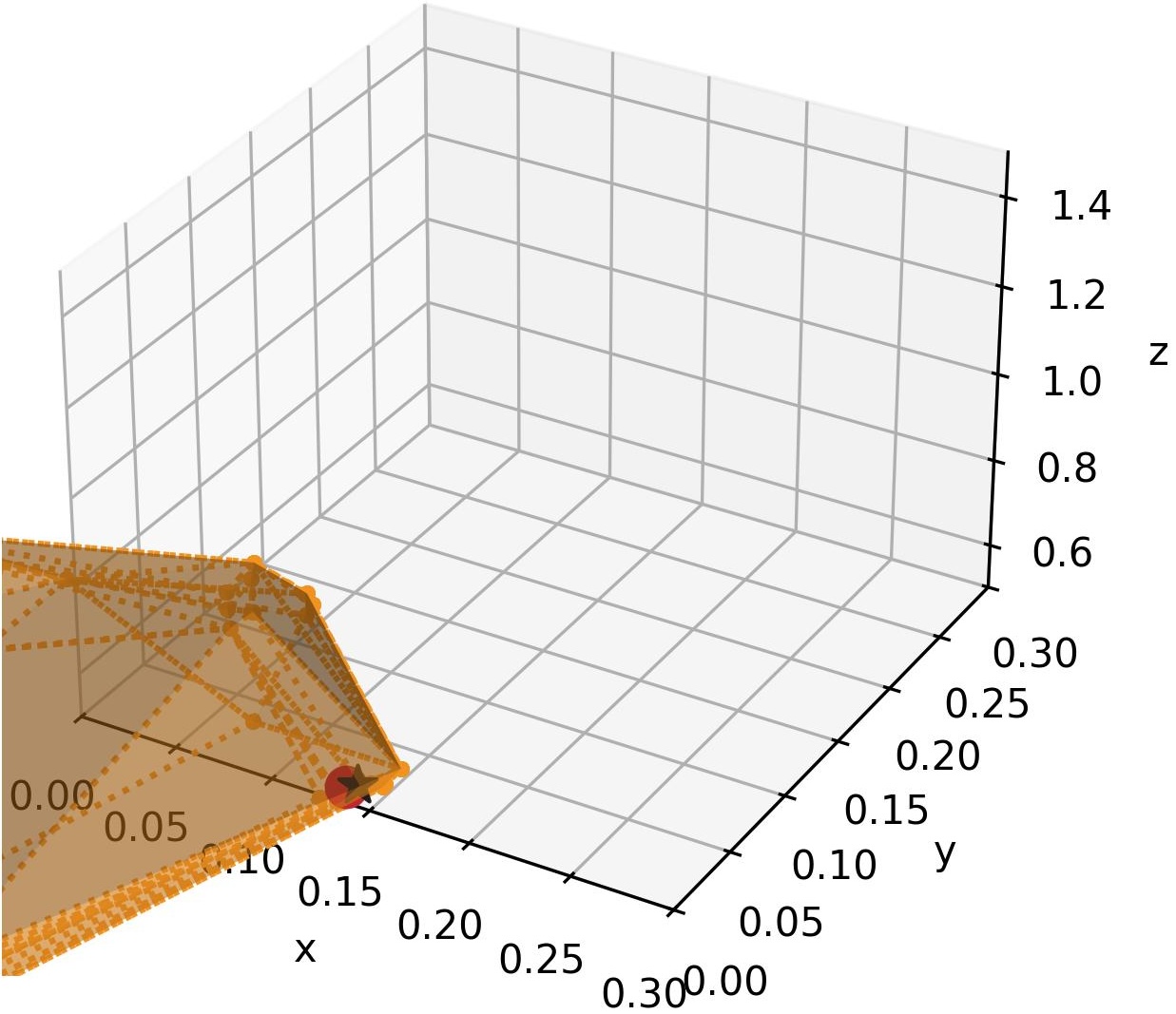}
        \caption{$\mathcal{T}^s_h:V_{\mathbf{t}}=14.8$}
    \end{subfigure}
    
    \caption{
    Visualization of the 6D confidence region for objects in LMO. 
    Our \( \mathcal{K}^r \) can cover the true keypoint within a user-defined \( \epsilon \), while the volumes of \( \mathcal{R}^d_r \) and \( \mathcal{T}^d_r \) are significantly smaller than that of \( \mathcal{R}^s_h \) and \( \mathcal{T}^s_h \).
    }
    \label{vis_sample_image_V_RT}
\end{figure*}

\begin{figure*}[htbp]
    \centering
    
    \begin{minipage}{0.24\textwidth}
        \centering
        \includegraphics[width=\textwidth]{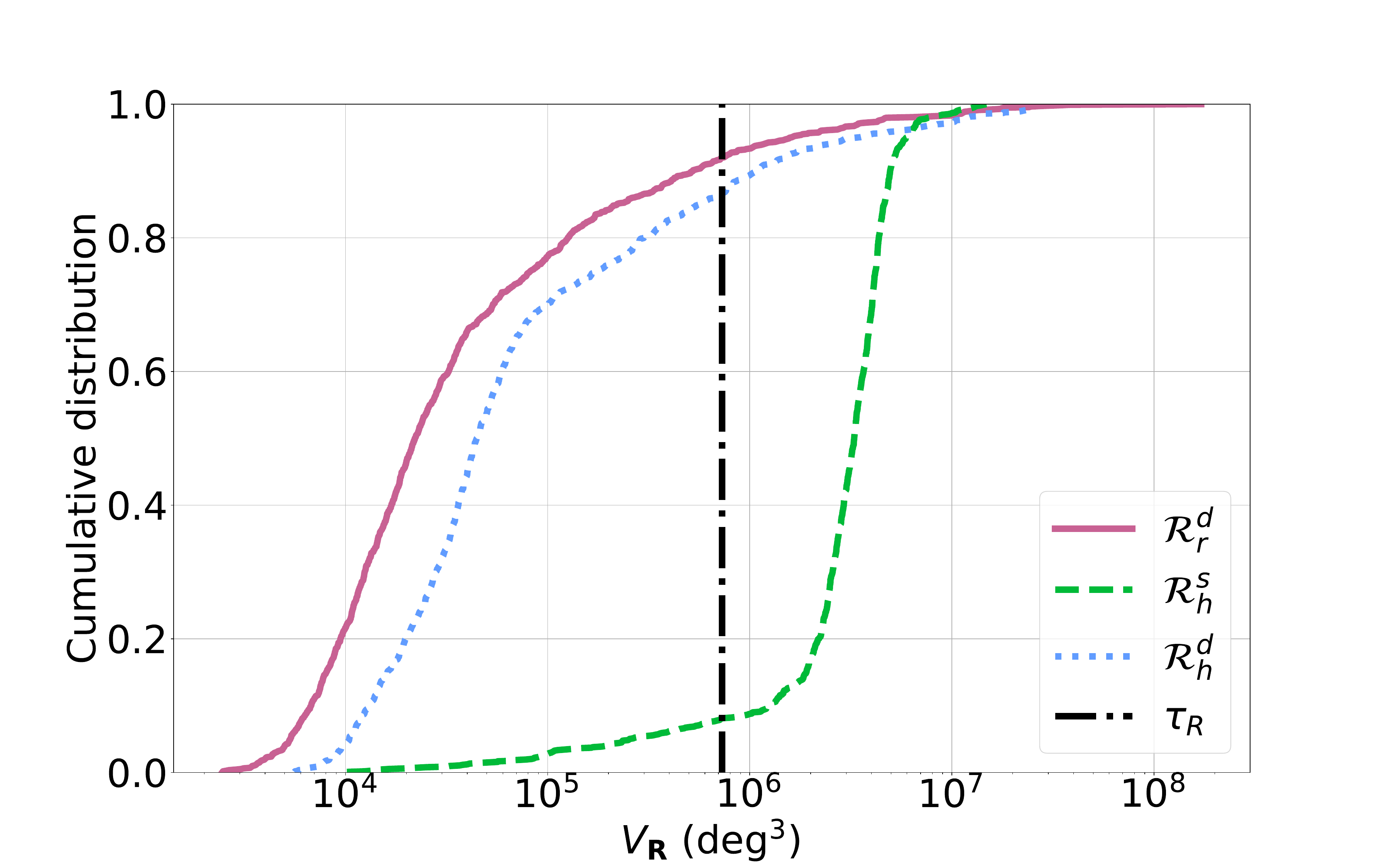}
        
    \end{minipage}
    \hfill
    \begin{minipage}{0.24\textwidth}
        \centering
        \includegraphics[width=\textwidth]{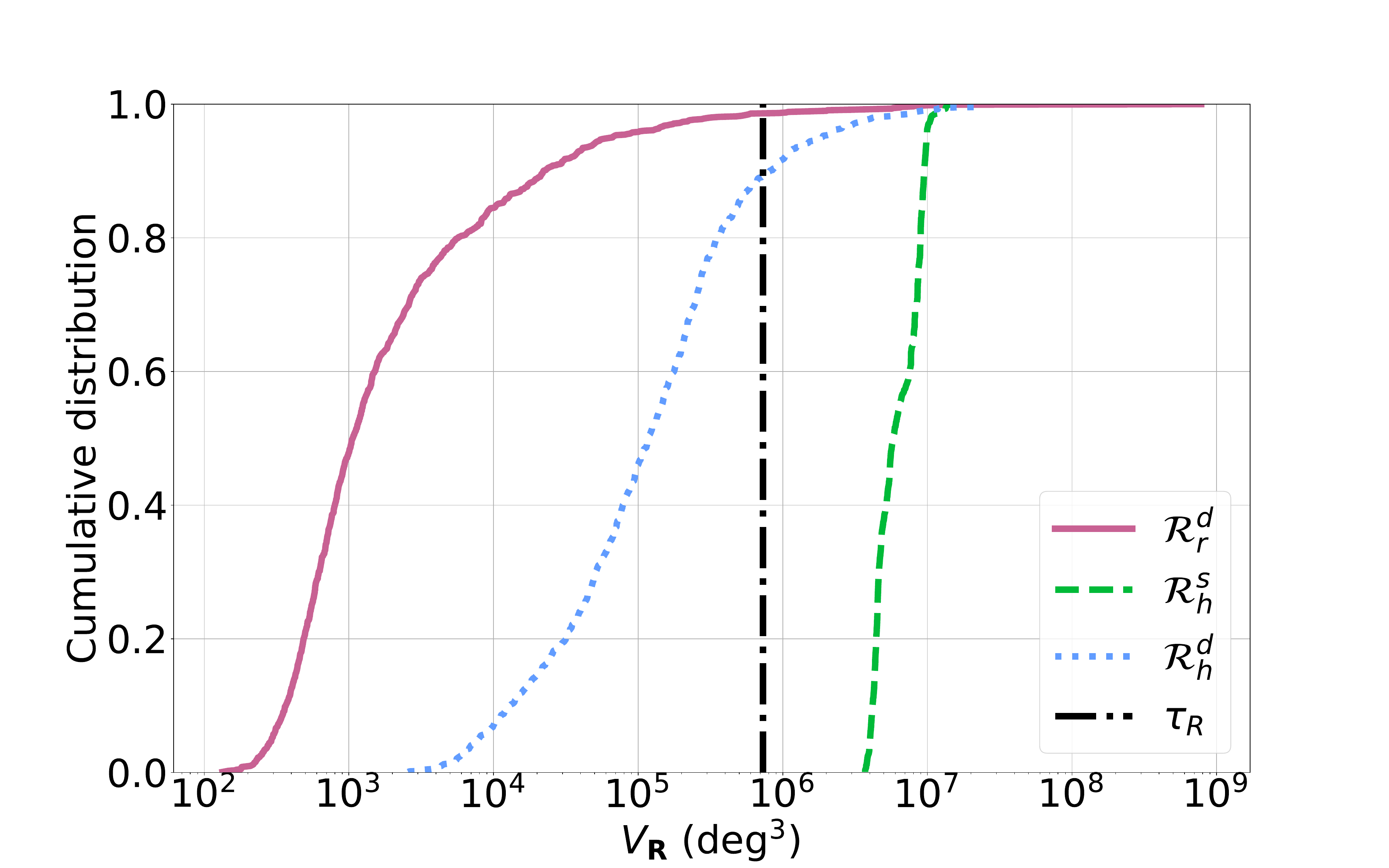}
        
    \end{minipage}
    \hfill
    \begin{minipage}{0.24\textwidth}
        \centering
        \includegraphics[width=\textwidth]{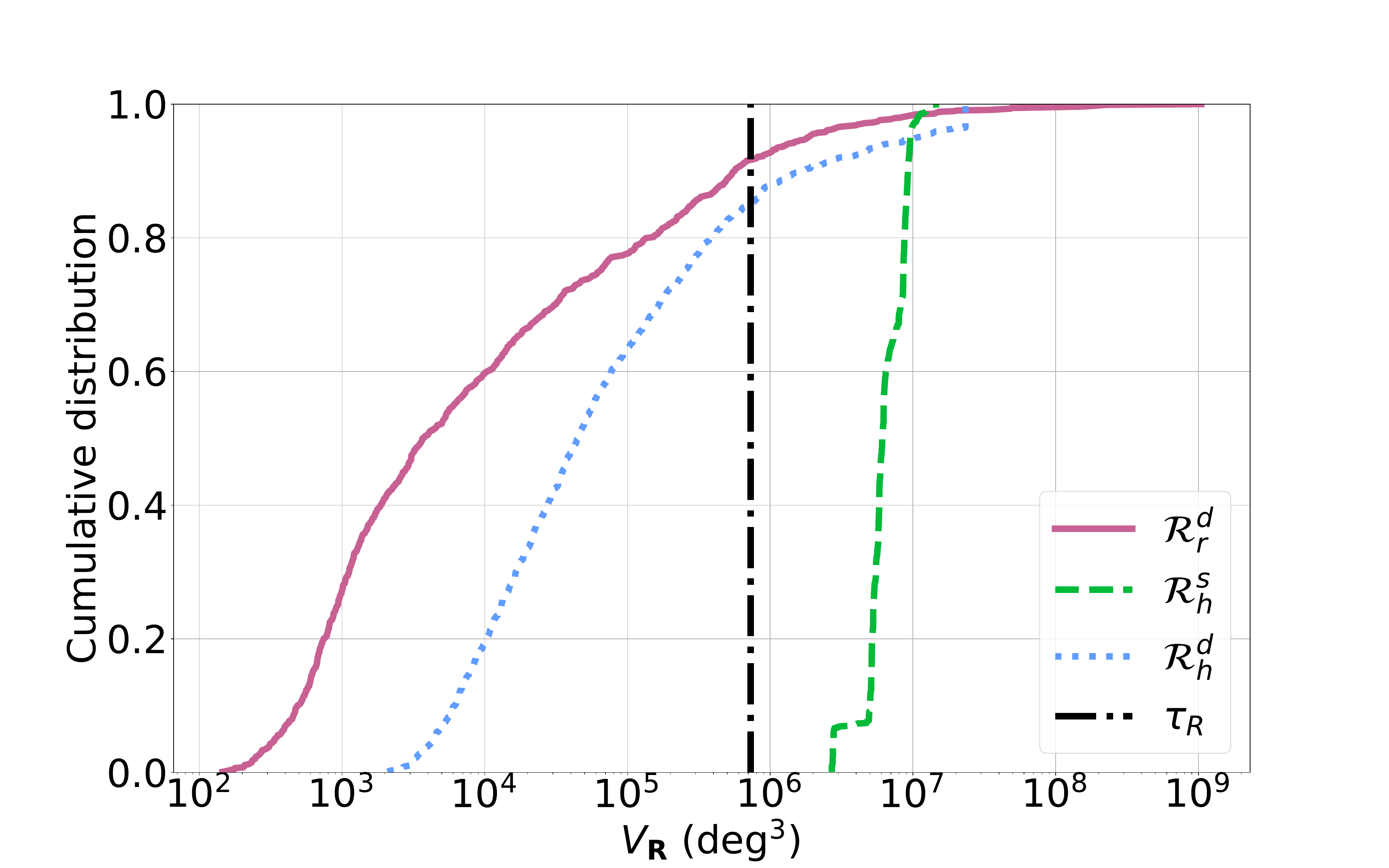}
        
    \end{minipage}
    \hfill
    \begin{minipage}{0.24\textwidth}
        \centering
        \includegraphics[width=\textwidth]{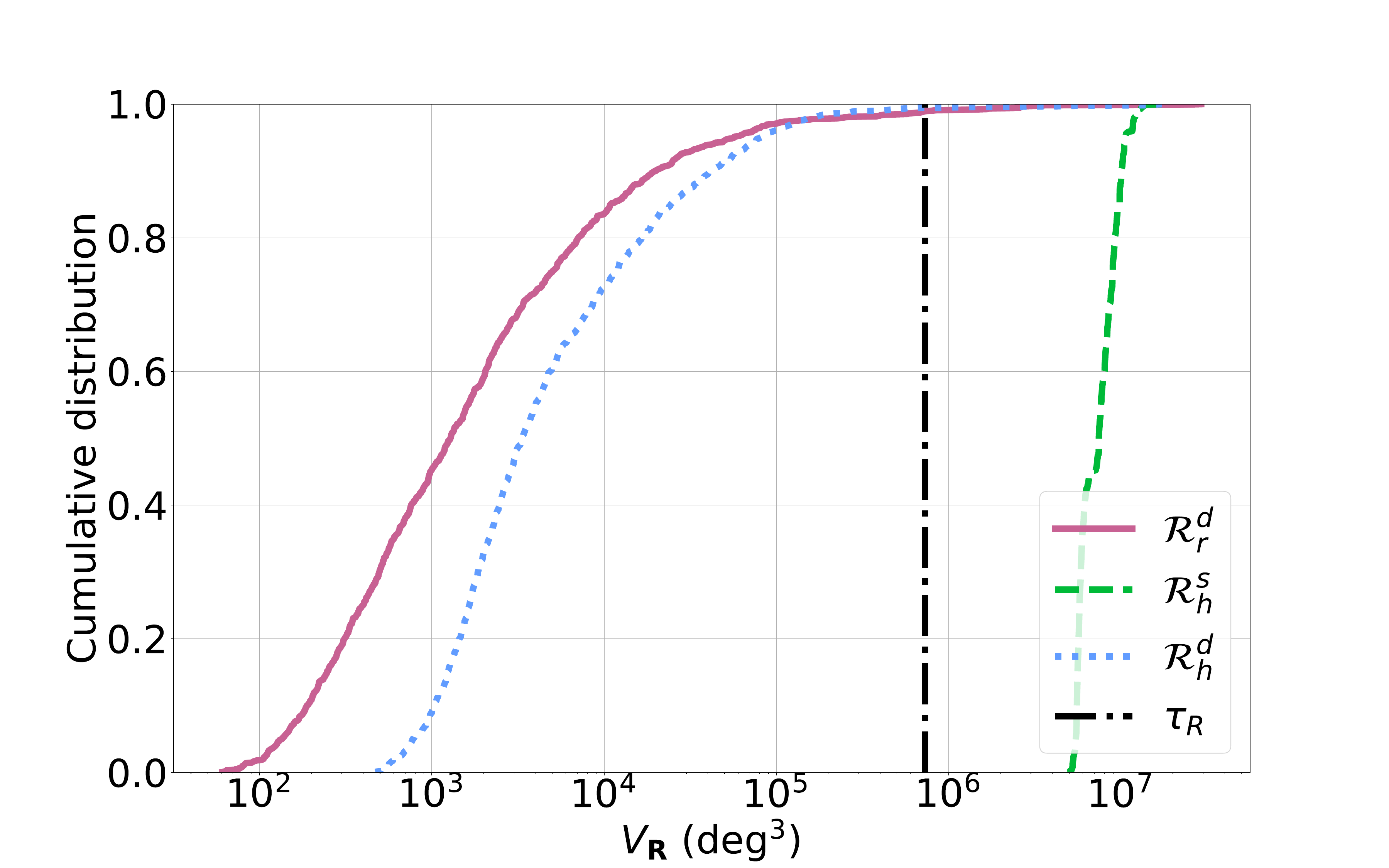}
        
    \end{minipage}
    \hfill 
    \begin{minipage}{0.24\textwidth}
        \centering
        \includegraphics[width=\textwidth]{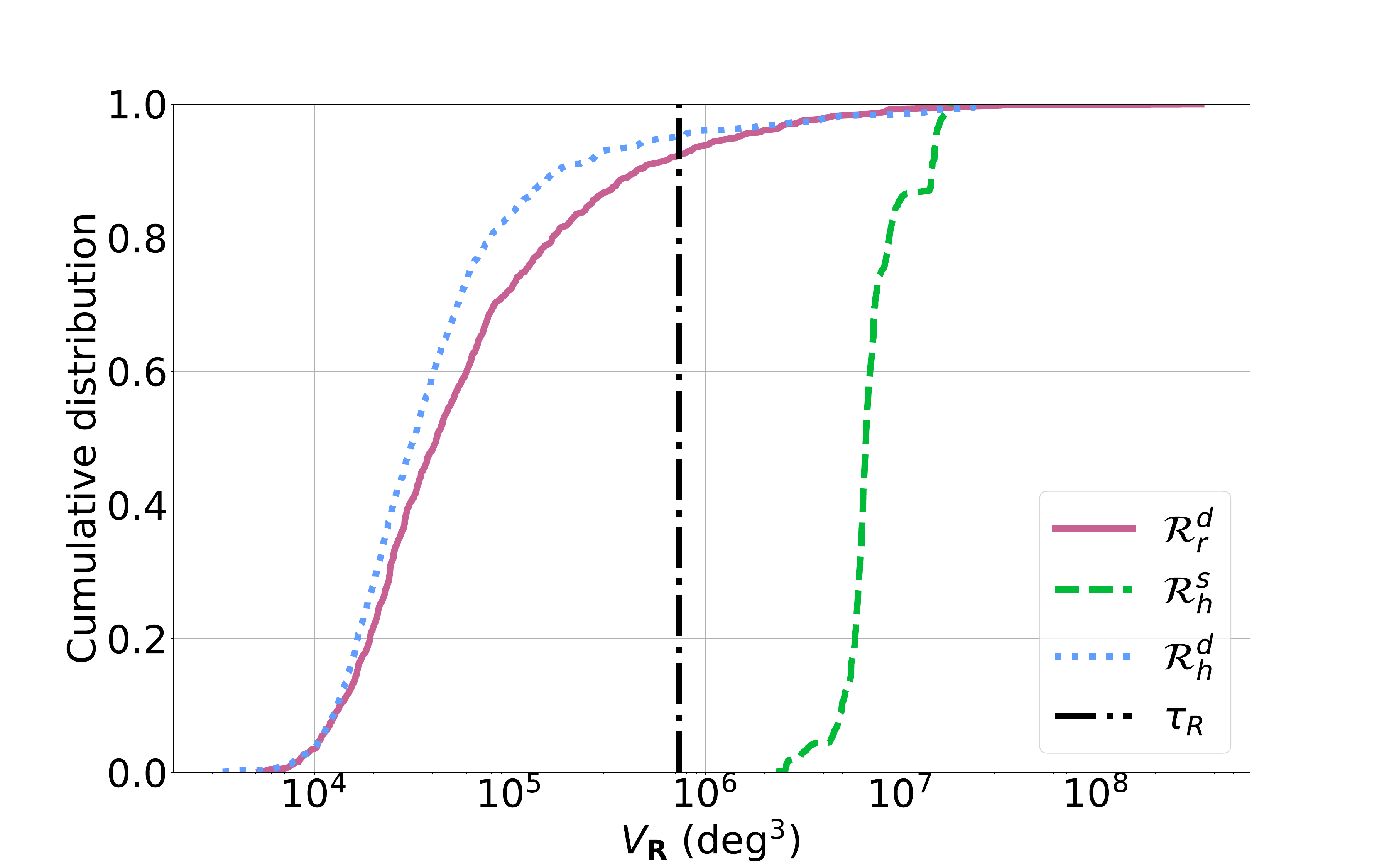}
        
    \end{minipage}
    \hfill
    \begin{minipage}{0.24\textwidth}
        \centering
        \includegraphics[width=\textwidth]{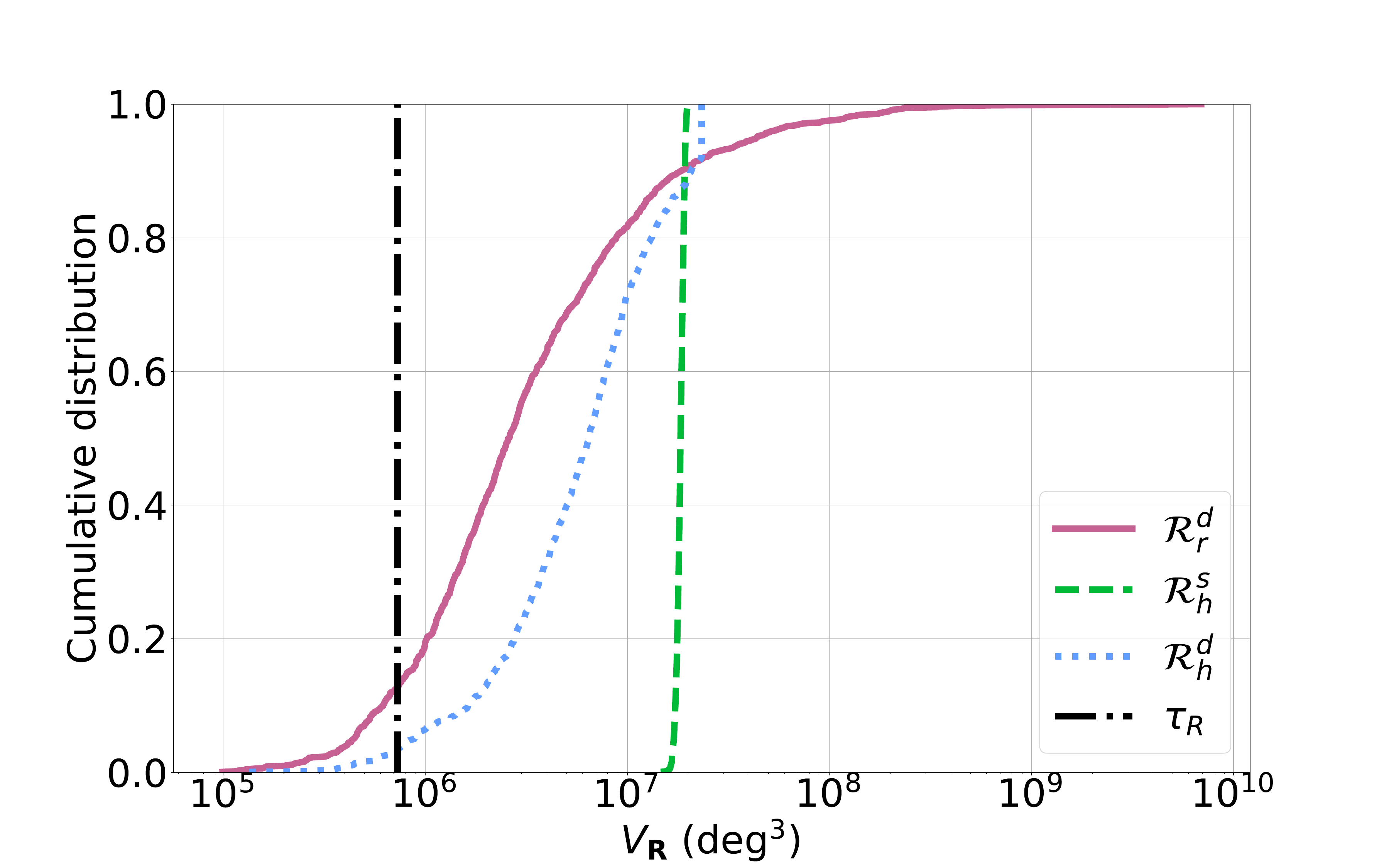}
        
    \end{minipage}
    \hfill
    \begin{minipage}{0.24\textwidth}
        \centering
        \includegraphics[width=\textwidth]{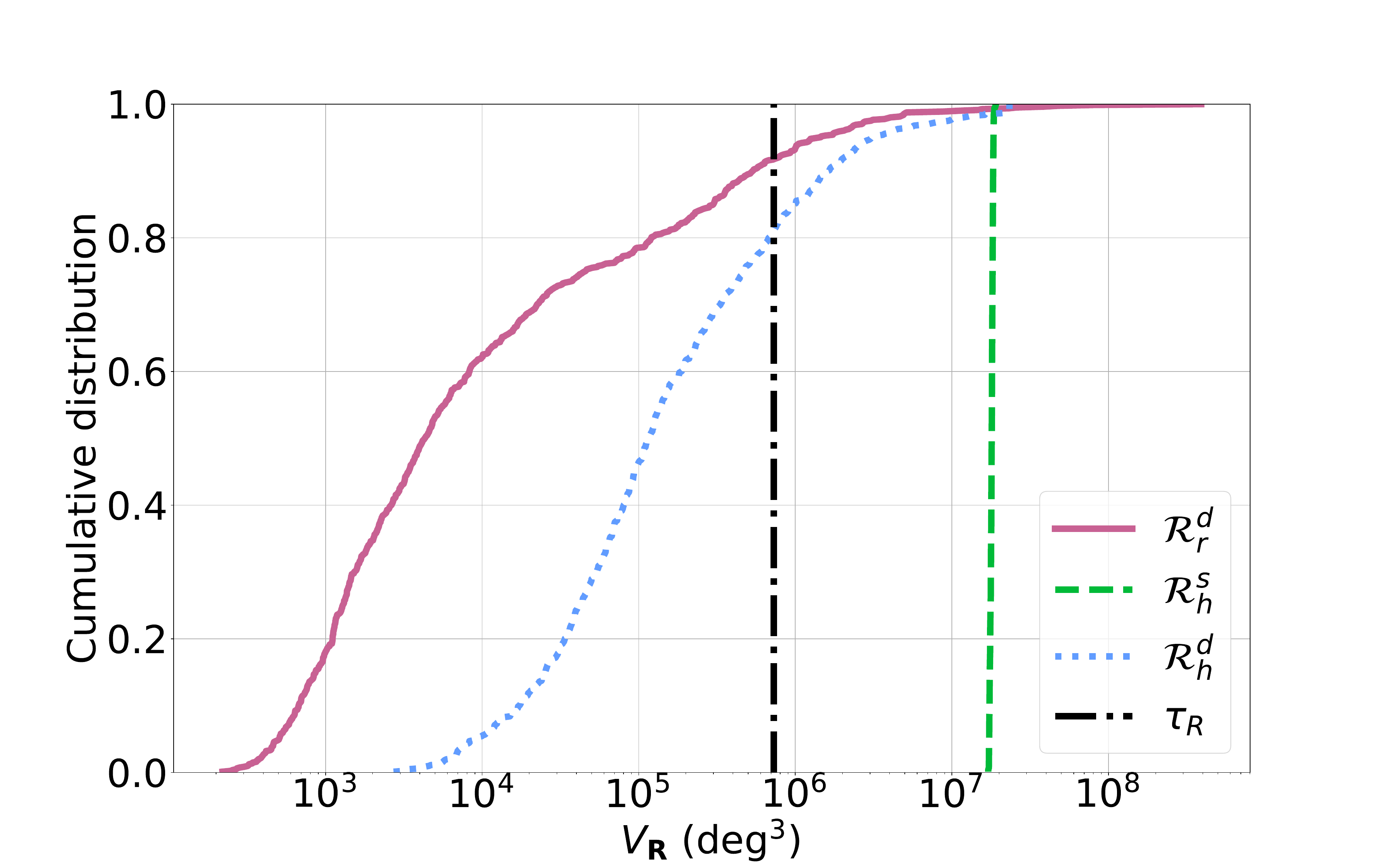}
        
    \end{minipage}
    \hfill
    \begin{minipage}{0.24\textwidth}
        \centering
        \includegraphics[width=\textwidth]{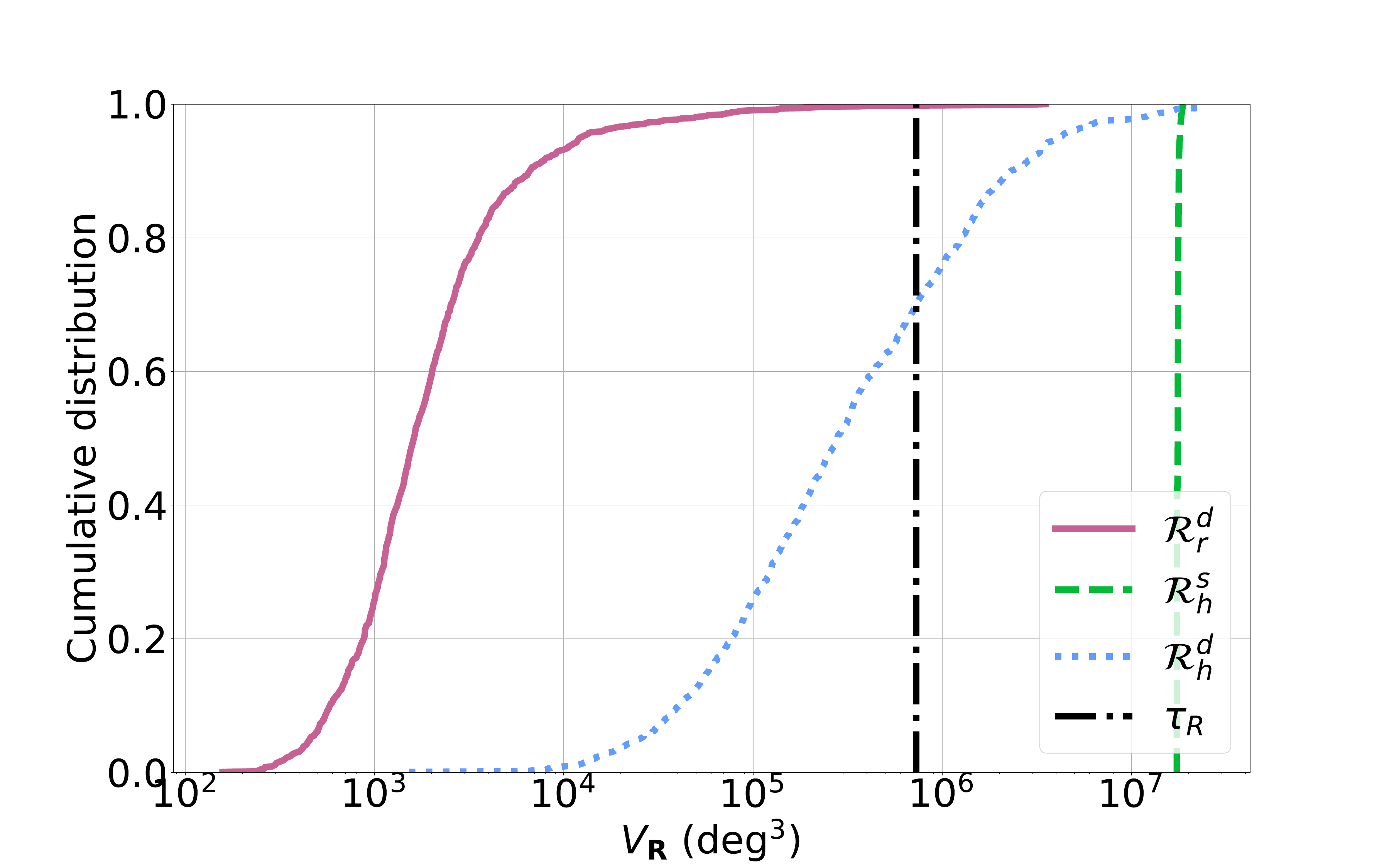}
        
    \end{minipage}
    
    \caption{Cumulative distribution function of the $V_{\mathbf{R}}$ with $\epsilon=0.1$ for the 8 objects in the LMO, the unit for $V_\mathbf{R}$ is \(10^3\ \deg^3\). 
    A CDF curve that is closer to the upper-left corner indicates a smaller \( V_{\mathbf{R}} \). The CDF curve of \( \mathcal{R}^d_r \) is closer to the upper-left corner than that of \( \mathcal{R}^s_h \).
    } 
    \label{fig_cdf_R}
\end{figure*}

 \textbf{6D pose confidence region volume:} 
However, if achieving a high coverage rate requires an excessively large confidence region, its practical utility may be greatly diminished.
Taking the Euler angles of rotation as an example, if the region spans $360^\circ \times  180^\circ \times  360^\circ$, it will inevitably contain any true Euler angle. 
Therefore, we also compute the volume of the pose confidence region as a metric.
The volume of the confidence ellipsoid \(\mathcal{C}^{d}_{r}\) is as follows:
\begin{equation}
V_\mathbf{R} = \frac{4}{3} \pi \sqrt{\det(\mathbf{\Sigma}_\mathbf{R})},
V_\mathbf{t} = \frac{4}{3} \pi \sqrt{\det(\mathbf{\Sigma}_\mathbf{t})}
\end{equation}
We set the volume threshold $\tau_{\mathbf{R}}=90^3\ \deg^3$ for $V_\mathbf{R}$, and $\tau_{\mathbf{t}}=1^3\ m^3$ for $V_\mathbf{t}$.
For sample-based \(\mathcal{C}^{s}_{h}\), 
its volume is given by the convex hull of the pose samples.
If the confidence region volume exceeds the specified threshold, 
it is considered out of bounds and excluded from the mean volume calculation. 
Additionally, when calculating pose coverage, it is regarded as a failure to cover the true pose.

\begin{table}[th]
\footnotesize
\centering
\begin{tabular}{c|cc|cc}
\hline
 & \multicolumn{2}{c|}{Ours} & \multicolumn{2}{c}{\cite{yang2023object}} \\ \cline{2-5} 
                       & \multicolumn{1}{l}{$\mathcal{K}^{r}$} & \multicolumn{1}{l|}{$\mathcal{C}^{d}_{r}$} & \multicolumn{1}{l}{$\mathcal{K}^{h}$} & \multicolumn{1}{l}{$\mathcal{C}^{s}_{h}$} \\ \hline
LMO~\cite{lmo_dataset_BrachmannKMGSR14}   & \textbf{0.0038} & \textbf{0.0361} & 0.0076 & 0.0550 \\ \hline
SPEED~\cite{Kisantal2019Satellite} & \textbf{0.0032} & \textbf{0.0358} & 0.0064 & 0.0521 \\ \hline

\end{tabular}
\caption{Time consumption for keypoint and pose confidence region estimation (unit: s).}
\label{tab_time}
\end{table}

\begin{table}[th]  
\footnotesize
\centering
\subcaptionbox{Acc of baseline methods and our approach\label{tab_2d_reproject}}[0.48\textwidth]{ 
\begin{tabular}{c|c|cc|c}
\hline
Acc & Ours & \multicolumn{2}{c|}{~\cite{yang2023object} $\mathcal{K}^{h}$} & ~\cite{peng2019pvnet} \\ 
\cline{3-4}
& $\mathcal{K}^{r}$ & $\epsilon=0.1$ & $\epsilon=0.4$ & \\ \hline
LMO mean          & $\textbf{74.45}$ & 67.33 & 70.71 & 61.06  \\ \hline
SPEED~\cite{Kisantal2019Satellite} & \textbf{97.09} & 57.80 & 57.40 & 57.46 \\ \hline
\end{tabular}%
}\hfil%
\subcaptionbox{$\eta^{kpt}$ of baseline methods and our approach\label{2d_point_coverage_rate}}[0.48\textwidth]{ 
\begin{tabular}{c|cl|ll}
\hline
$\eta^{kpt}$ & \multicolumn{2}{c|}{Ours $\mathcal{K}^{r}$} & \multicolumn{2}{c}{\cite{yang2023object} $\mathcal{K}^{h}$} \\ 
\cline{2-5} 
& $\epsilon=0.1$ & $\epsilon=0.4$ & $\epsilon=0.1$ & $\epsilon=0.4$ \\ \hline
LMO mean          & 90.65 & 60.38 & \textbf{91.16} & 63.38 \\ \hline
SPEED~\cite{Kisantal2019Satellite} & \textbf{89.66} & 61.25 & 88.88 & 62.64 \\ \hline
\end{tabular}%
}
\caption{Comparison of 2D keypoint confidence region results} 
\label{tab:total_2d_confidence_region}
\end{table}

\begin{table}
\centering
\footnotesize
\begin{tabular}{c|cc|cc|cc}
\hline
LMO & \multicolumn{2}{c|}{Ours}                                 & \multicolumn{2}{c|}{\cite{yang2023object} + Samp.}                  & \multicolumn{2}{c}{\cite{yang2023object} + Det.}                     \\ \cline{2-7} 
Objects & $\mathcal{T}^{d}_{r}$   & $\mathcal{R}^{d}_{r}$  & $\mathcal{T}^{s}_{h}$  & $\mathcal{R}^{s}_{h}$   & $\mathcal{T}^{d}_{h}$      & $\mathcal{R}^{d}_{h}$\\ \hline
1  &  70.52   &   91.26  & \textbf{97.26} &  N/A & 76.88  &  \textbf{93.38} \\
2  &  88.73   &   89.98  &  99.25  &  N/A & \textbf{99.59}  &  \textbf{98.18} \\
3  &  77.28   &   87.55  & \textbf{98.29}   &  N/A & 88.02  &  \textbf{90.59} \\
4  &  85.09   &   96.62  &  \textbf{97.12}  & N/A & 90.28 &  \textbf{98.85} \\
5  &  97.18   &   \textbf{79.70}  &  \textbf{99.81} & N/A  & 90.13 &  76.41\\
6  &  98.36   &   \textbf{1.46}  &  77.81 & N/A  &  \textbf{98.63} &  1.37 \\
7  &  79.73   &   87.76  & \textbf{98.64}  & N/A  &  89.99 & \textbf{91.84} \\
8  &  69.01   &   86.94  & \textbf{99.92}  & N/A  &  98.51 & \textbf{98.02} \\ \hline
mean & 83.24  &   77.66  & \textbf{96.61} & N/A & 91.50 & \textbf{81.08} \\ \hline
SPEED & 86.69 & 88.81 & 6.40 & N/A & \textbf{87.10} & \textbf{90.92} \\ \hline
\end{tabular}
\caption{$\eta^{\mathbf{R}}$ and $\eta^{\mathbf{t}}$ of baselines and our approach with $\epsilon=0.1$. 
‘N/A’ represents all regions' volumes overflow the threshold.}
\label{tab_6d_cover_rate}
\end{table}

\begin{figure*}[htbp]
    \centering
    \includegraphics[width=0.95\textwidth]{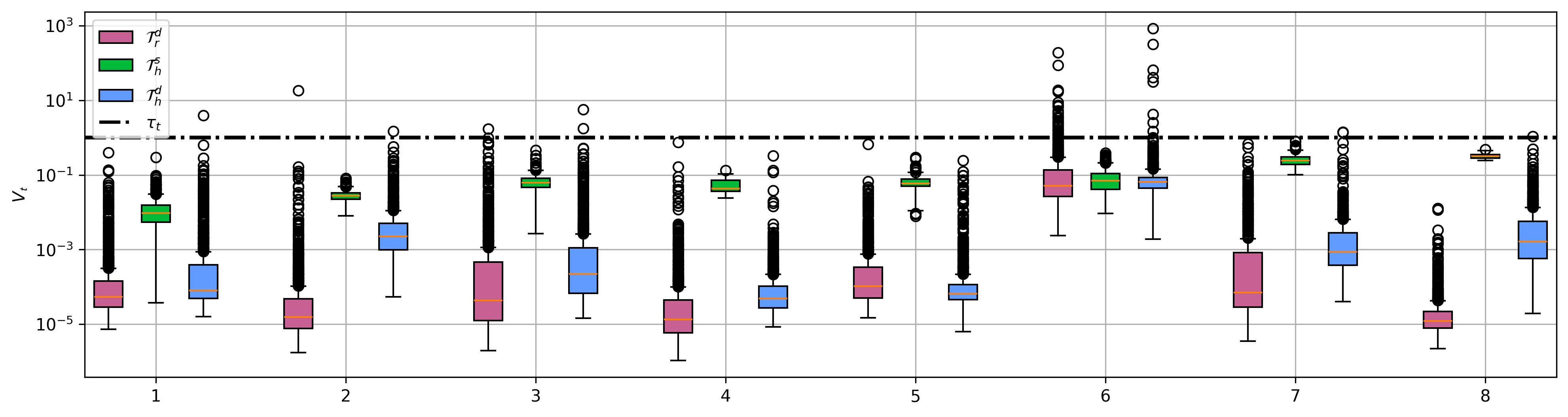}  
    \caption{A boxplot analysis of baseline and proposed method for $V_{\mathbf{t}}$ on the LMO dataset. 
    The boxplot for $\mathcal{T}^d_r$ exhibits consistently lower values across all key statistical measures compared to $\mathcal{T}^s_h$.
    After applying deterministic regression ($\mathcal{T}^d_h$), $V_{\mathbf{t}}$ also decreases significantly.
    }
    \label{fig:volume_box_T}
\end{figure*}

\begin{table*}
  \centering
  \footnotesize
  \begin{tabular}{c|cccc|cccc|cccc}
  \hline
  LMO & \multicolumn{4}{c|}{Ours}                                 & \multicolumn{4}{c|}{\cite{yang2023object} + Samp.}                  & \multicolumn{4}{c}{\cite{yang2023object} + Det.}                     \\ \cline{2-13}

  Objects & $\mathcal{T}^{d}_{r}$ & Out   & $\mathcal{R}^{d}_{r}$ &Out  & $\mathcal{T}^{s}_{h}$ &Out  & $\mathcal{R}^{s}_{h}$ &Out   & $\mathcal{T}^{d}_{h}$ &Out      & $\mathcal{R}^{d}_{h}$ &Out\\ \hline
  1    & 0.9 & 0 & \textbf{28.96} & 69 & 12.6 & 0 & 252.2 & 1041 &   \textbf{0.5}  & 0 & 50.3 & 64  \\
  2    & \textbf{0.7} & 0 & \textbf{9.9} & 9 & 28.3 & 0 & N/A & 1207   & 0.9 & 0 & 45.0 & 12  \\
  3    & 2.8 & 3 & 50.7 & 113 & 66.8 & 0 & N/A & 1052   & \textbf{1.4} & 0 & \textbf{42.9} & 73  \\
  4    & 1.5 & 0 & 6.5 & 12  & 52.0 & 0 & N/A & 1214   & \textbf{0.1} & 0 & \textbf{3.5} & 2   \\
  5    & 0.4 & 0 & 52.8 & 43 & 63.1 & 0 & N/A & 1064   & \textbf{0.2} & 0 & \textbf{24.7} & 24  \\
  6    & \textbf{57.6} & 11 & \textbf{514.4} & 1055& 82.2 & 1 & N/A & 1095   &  59.9 & 10 & 515.7 & 1057 \\
  7    & 4.8 & 0 & \textbf{62.9} & 98  & 252.4 & 0 & N/A & 809   &  \textbf{0.8}  & 0 & 67.5 & 56 \\
  8    & \textbf{0.03} & 0 & \textbf{4.6} & 3   & 322.2 & 0 & N/A & 1210  &  0.3  & 0 & 51.6 & 9 \\ \hline
  mean & 8.6 & 2 & \textbf{91.3} &  175 & 109.9 & 0 & 252.2 & 1070 & \textbf{8.0}  & 1 & 100.1  & 162  \\ \hline
  SPEED& \textbf{0.7} & 0 & \textbf{0.2} & 0 & 287.8 & 920 & 210.6 & 898& 73.4 & 50 & 4.4 & 18 \\ \hline

  \end{tabular}
  \caption{$V_{\mathbf{t}}$ and $V_{\mathbf{R}}$ with $\epsilon=0.1$: The unit for the rotational and translational confidence region are \(10^3\ \deg^3\) and \(10^{-3}\ m^3\). 'N/A' indicates that all confidence regions exceed the threshold, and 'Out' refers to the number of images where the threshold is exceeded.}
  \label{6d_pose_volume}
  \end{table*}
  
\subsection{Experimental Results}
\textbf{Accuracy and speed:}
We provide a concise comparison of $\text{Acc}$ with baseline methods in \cref{tab:total_2d_confidence_region} (a). While prior methods such as \cite{yang2023object} rely on heatmap calibration and occlusion-aware sampling to improve accuracy in challenging scenarios, their improvements diminish in occlusion-free environments (e.g., on SPEED \cite{Kisantal2019Satellite}). 
Our method universally surpasses \cite{yang2023object,peng2019pvnet} across both cases with various $\epsilon$. 
The comprehensive results for the 8 LMO objects are presented in supplementary~\cref{tab_2d_reproject}.

For heatmap-based keypoint detection and sampling-based confidence region prediction, increasing the size of heatmaps and the number of samples significantly raises inference and sampling time. 
As shown in \cref{tab_time}, our regression-based keypoint detection and deterministic confidence region prediction reduce time consumption by 50\% and 31.3\% on SPEED, and by 50\% and 34.4\% on LMO, compared to \cite{yang2023object}. 
In \cite{yang2023object}, Faster R-CNN and the Stacked Hourglass are used for bounding box detection and keypoint detection, respectively. 
For a fair comparison, we use the same bounding box results as \cite{yang2023object}. 
The Stacked Hourglass has 26.4 million parameters and 26.77 GFlops, while our method has only 11.1 million parameters and 7.19 GFlops.

\textbf{2D Keypoint Confidence Region:}
After applying ICP, the confidence region coverage rates of $\mathcal{K}^{h}$ and $\mathcal{K}^{r}$ closely align with $1-\epsilon$ in both SPEED and LMO datasets, as shown in \cref{tab:total_2d_confidence_region} (b). 
The comprehensive results for the each object in the LMO dataset  are presented in supplementary~\cref{2d_point_coverage_rate}.
However, the 2D keypoint coverage can not fully reflect the 6D pose coverage. 
Therefore, we proceed to evaluate the confidence region coverage and volume for the 6D pose.

\textbf{6D Pose Confidence Region Coverage Rates:}
To compare the sampling-based method with our deterministic approach, 
\cref{tab_6d_cover_rate} presents the 6D pose coverage rates for three pose confidence regions: 
regions obtained by direct propagation ($\mathcal{C}^{d}_{h}$) and sampling ($\mathcal{C}^{s}_{h}$) from $\mathcal{K}^{h}$ generated by \cite{yang2023object},
and our deterministically propagated $\mathcal{C}^{d}_{r}$ from $\mathcal{K}^{h}$. 
Our single-shot PnP algorithm for pose estimation (\cref{sec:single_shot_pnp}) does not utilize sampling or the RANSAC method. 
For translation in the LMO dataset, 
although $\mathcal{T}^{s}_{h}$ achieves a higher coverage rate compared to $\mathcal{T}^{d}_{r}$ (96.61\% vs. 83.24\%), 
its volume is significantly larger, 
as shown in \cref{6d_pose_volume} (109.9 $dm^3$ vs. 8.6 $dm^3$). 
On the SPEED dataset, the larger range of translation causes more $\mathcal{T}^{s}_{h}$ volumes to exceed the threshold, 
thereby lowering its pose coverage rate.

\textbf{6D Pose Confidence Region Volume:}
As shown in \cref{6d_pose_volume}, replacing the original sampling method in \cite{yang2023object} with our deterministic propagation method results in $\mathcal{R}^{d}_{h}$ volumes that fall below the threshold while still meeting coverage requirements. When comparing our deterministic keypoint regression approach (1st column of \cref{6d_pose_volume}) with the heatmap-based approach in \cite{yang2023object} (3rd column of \cref{6d_pose_volume}), heatmaps better accommodate occlusions in the LMO dataset, achieving confidence region volumes comparable to ours. However, heatmaps prove unsuitable for the occlusion-free SPEED dataset, leading to a significant increase in $V_{\mathbf{R}}$ and $V_{\mathbf{t}}$. 
As demonstrated in \cref{vis_sample_image_V_RT}, our method achieves a statistically significant reduction in both $V_\mathbf{R}$ and $V_\mathbf{t}$ compared to \cite{yang2023object}. 
Additional visualizations can be found in the supplementary material.
The largest pose confidence region volume is observed with $\mathcal{C}^{s}_{h}$ (2nd col. of \cref{6d_pose_volume}), where most $\mathcal{R}^{s}_{h}$ volumes exceed the threshold. The mean volume of $\mathcal{T}^{s}_h$ remains within the threshold but is still the highest, indicating that sampling-based confidence regions fail to cover the ground truth pose within a narrow range.

Transitioning from $\mathcal{C}^{s}_{h}$ to $\mathcal{C}^{d}_{r}$ reduces $V_{\mathbf{R}}$ by 63.8\% and 99.9\%, and $V_{\mathbf{t}}$ by 92.2\% and 99.8\% across the two datasets. 
In the occlusion-free SPEED, our proposed method reduces the volumes of $\mathcal{T}^{d}_{r}$ and $\mathcal{R}^{d}_{r}$ to as low as $0.7 \times 10^{-3}\ m^3$ and $0.2 \times 10^3\ \deg^3$. 
To elucidate the statistical disparities in region volumes across diverse methods, we visualize the CDF curves of $V_{\mathbf{R}}$ (\cref{fig_cdf_R}) and the boxplot of $V_{\mathbf{t}}$ (\cref{fig:volume_box_T}) within the LMO. 
In the CDF curve, $\mathcal{R}^{d}_{r}$ is closer to the top-left corner, while most $\mathcal{R}^{s}_{h}$ exceeds $\tau_{\mathbf{R}}$, validating the efficacy of our deterministic propagation method in reducing region size. 
Furthermore, as shown in \cref{fig:volume_box_T}, $V_{\mathbf{t}}$ of $\mathcal{T}^{d}_{r}$ is reduced by several orders of magnitude compared to $\mathcal{T}^{s}_{h}$.

\section{Conclusions}
Our framework introduces an efficient pipeline for estimating compact 6D pose confidence regions that achieve real-time operational capability. 
Crucially, we develop novel evaluation metrics that consider both ground-truth coverage probability and volumetric compactness of the confidence regions. 
Benchmark evaluations on the LineMOD Occlusion and SPEED datasets demonstrate significant improvements: 
1) 33\% faster inference speeds compared to the baseline, 
2) up to 99\% decrease in confidence region volume, 
while maintaining comparable coverage rates across test scenarios. 
But our Gaussian pose assumption is a key limitation, 
as the non-linear 2D-to-6D transformation challenges its coverage guarantees, 
which we will address by exploring more suitable distributions.

{
    \small
    \bibliographystyle{unsrt}
    \bibliography{main}
}

\onecolumn

\section{Supplementary Experiments}
\label{sec:supplementary}
\subsection{Keypoint Confidence Regions}
\label{sec:supp_Keypoint_confidence_regions}

\begin{table*}[h]  
\centering
\begin{minipage}{0.48\textwidth}  
\centering
\begin{tabular}{c|c|cc|c}
\hline
LMO~\cite{lmo_dataset_BrachmannKMGSR14} & Ours & \multicolumn{2}{c|}{~\cite{yang2023object}}       & ~\cite{peng2019pvnet} \\ \cline{3-4}
Objects &  & $\epsilon=0.1$ & $\epsilon=0.4$ &                       \\ \hline
ape (1)                      & 76.78 & 77.70 & $\textbf{79.52}$ & 69.14 \\
can (2)                     & $\textbf{91.96}$ & 73.41 & 75.97 & 86.09 \\
cat (3)                     & 90.11 & 87.36 & $\textbf {90.59}$ & 65.12                 \\
driller (4)                 & $\textbf{89.29}$ & 79.32 & 83.08 & 61.44 \\
duck (5)                    & $\textbf{84.39}$ & 82.71 & 82.54 & 73.06 \\
eggbox (6)                  & 6.85 & 0 & 0 & $\textbf{8.43}$ \\
glue (7)                    & 69.83 & 56.49 & $\textbf{71.08}$ & 55.37  \\
holepuncher (8)             & $\textbf{86.44}$ & 81.65 & 82.89 & 69.84  \\ \hline
mean                     & $\textbf{74.45}$ & 67.33 & 70.71 & 61.06  \\ \hline
SPEED~\cite{Kisantal2019Satellite} & \textbf{97.09} & 57.80  & 57.40 &  57.46 \\ \hline
\end{tabular}
\caption{Acc of baseline methods and our approach}
\label{tab_2d_reproject}

\end{minipage}
\hfill  
\begin{minipage}{0.48\textwidth}  
\centering
\begin{tabular}{c|cl|ll}
\hline
LMO~\cite{lmo_dataset_BrachmannKMGSR14} & \multicolumn{2}{c|}{Ours\ $\mathcal{K}^{r}$}       & \multicolumn{2}{c}{\cite{yang2023object}\ $\mathcal{K}^{h}$}              \\ \cline{2-5} 
Objects &$\epsilon=0.1$ &$\epsilon=0.4$& $\epsilon=0.1$ & $\epsilon=0.4$ \\ \hline
1  & \textbf{90.37}  &  61.51  & 88.35  & 64.87  \\
2  & 89.97 &  59.90 & \textbf{93.54}  & 61.81   \\
3  & 91.63 & 64.63 & \textbf{92.30}  & 63.69   \\
4 & \textbf{91.84} & 59.30  & 90.86  & 64.74   \\
5 & \textbf{90.31} & 64.09 & 90.13  & 64.94   \\
6 &  88.40 & 59.90 & \textbf{88.86}  & 60.37   \\
7   &  \textbf{92.02}  & 54.34 & 91.97  & 59.83   \\
8 & 90.66 & 59.42  & \textbf{93.31}  & 66.86   \\ \hline
mean  & 90.65  & 60.38 & \textbf{91.16}  &63.38   \\ \hline
SPEED~\cite{Kisantal2019Satellite} & \textbf{89.66} & 61.25 & 88.88 & 62.64   \\ \hline
\end{tabular}
\caption{$\eta^{kpt}$ of baseline methods and our approach}
\label{2d_point_coverage_rate}
\end{minipage}
\end{table*}

\subsection{Qualitative Pose Confidence Regions}
\label{sec:supp_pose_confidence_regions}

\begin{figure*}[h]
    \centering
    \includegraphics[width=1\textwidth]{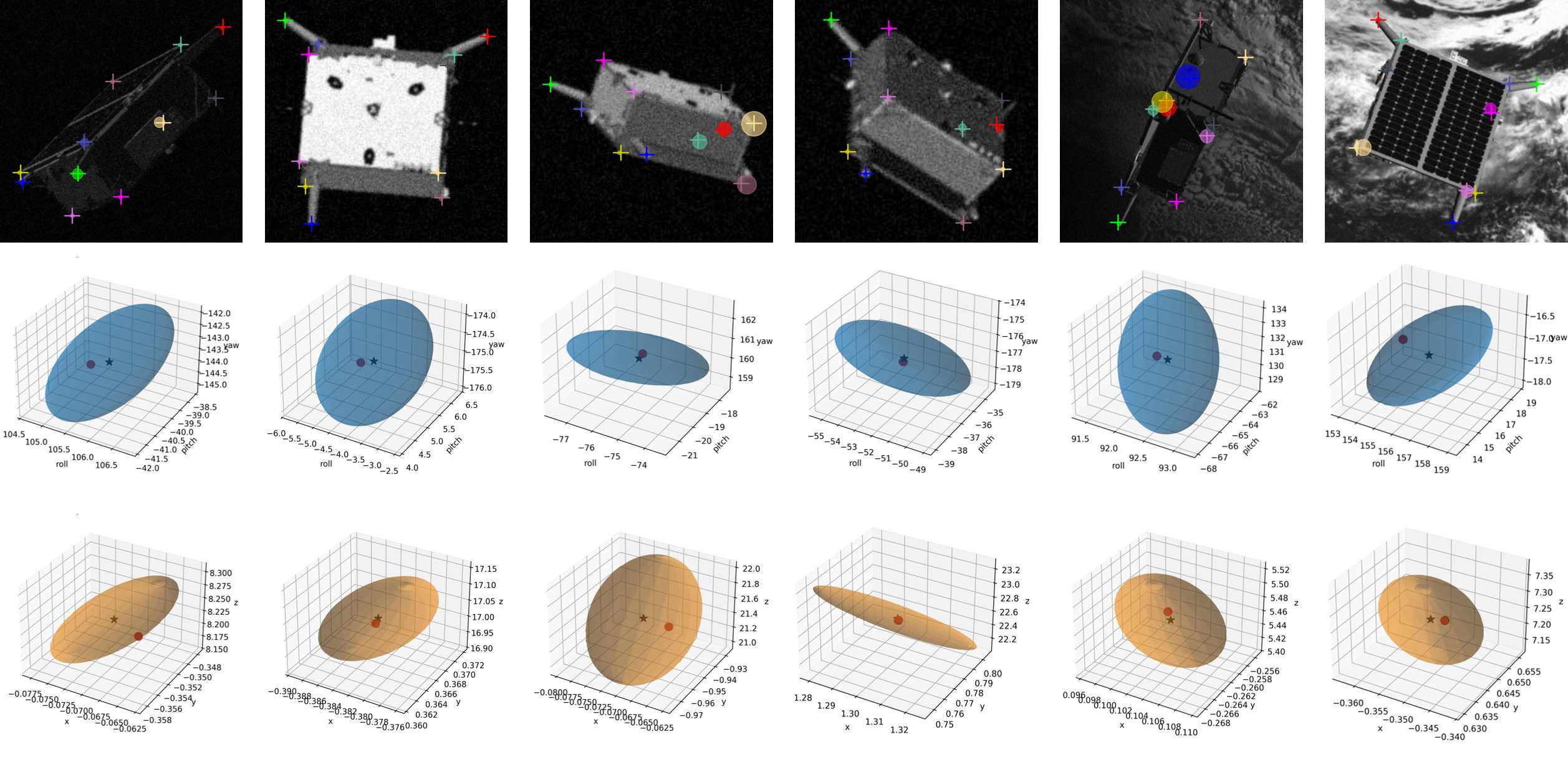}  
    \caption{$\mathcal{K}^{r}$, $\mathcal{R}^{d}_{r}$, and $\mathcal{T}^{d}_{r}$ (corresponding to rows 1 to 3) for six images (corresponding to columns 1 to 6) in the SPEED dataset when $\epsilon = 0.1$.}
    \label{fig_eps_10_speed}
\end{figure*}

\begin{figure*}[ht]
    \centering
    \includegraphics[width=1\textwidth]{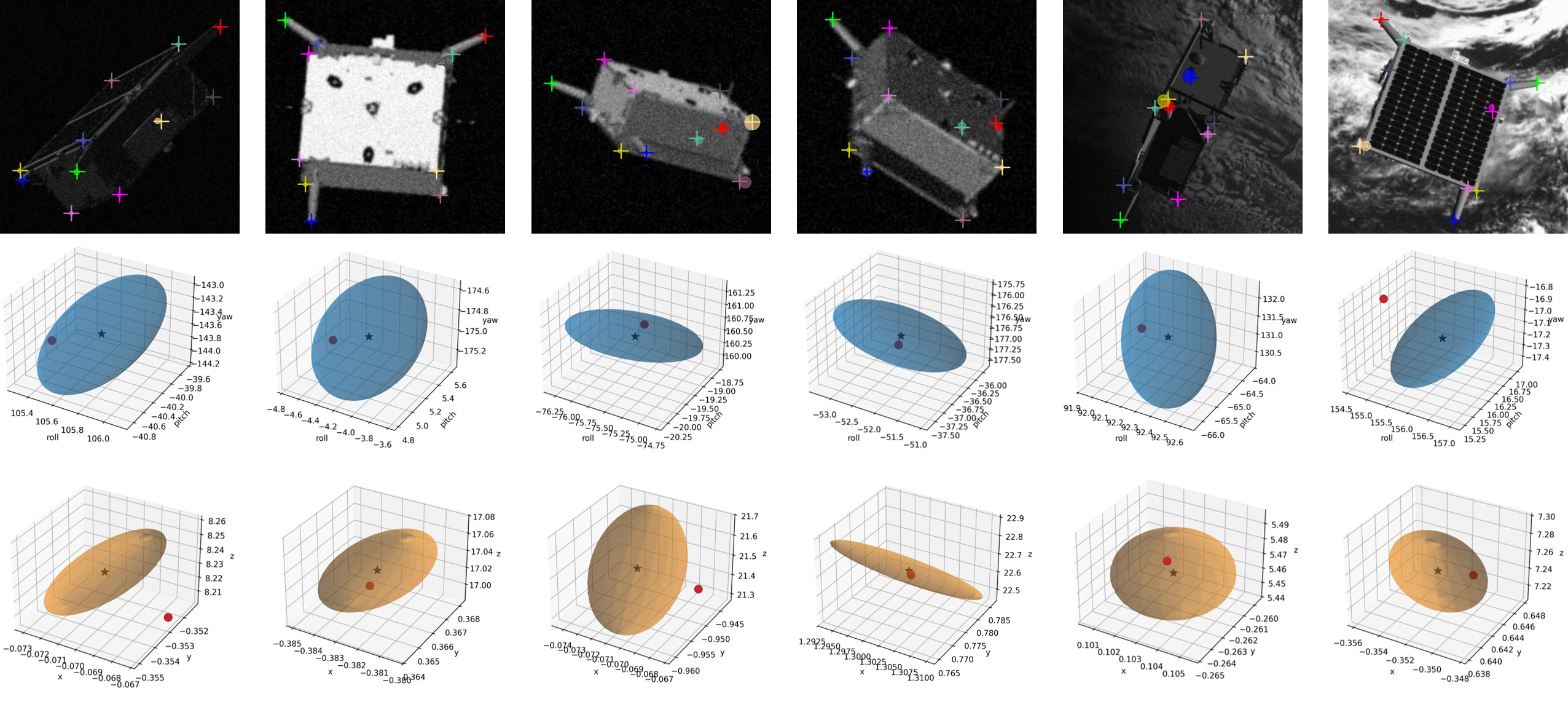}  
    \caption{$\mathcal{K}^{r}$, $\mathcal{R}^{d}_{r}$, and $\mathcal{T}^{d}_r$ (corresponding to rows 1 to 3) for six images (corresponding to columns 1 to 6) in the SPEED dataset when $\epsilon = 0.4$.}
    \label{fig_eps_40_speed}
\end{figure*}

Fig.~\ref{fig_eps_10_speed} and Fig.~\ref{fig_eps_40_speed} provides SPEED satellites’ $\mathcal{K}^{r}$, $\mathcal{R}^{d}_{r}$, and $\mathcal{T}^{d}_r$ with $\epsilon=0.1$ and $\epsilon=0.4$ across varying viewpoints and diverse backgrounds.


\begin{figure*}[h]
    \includegraphics[width=\textwidth]{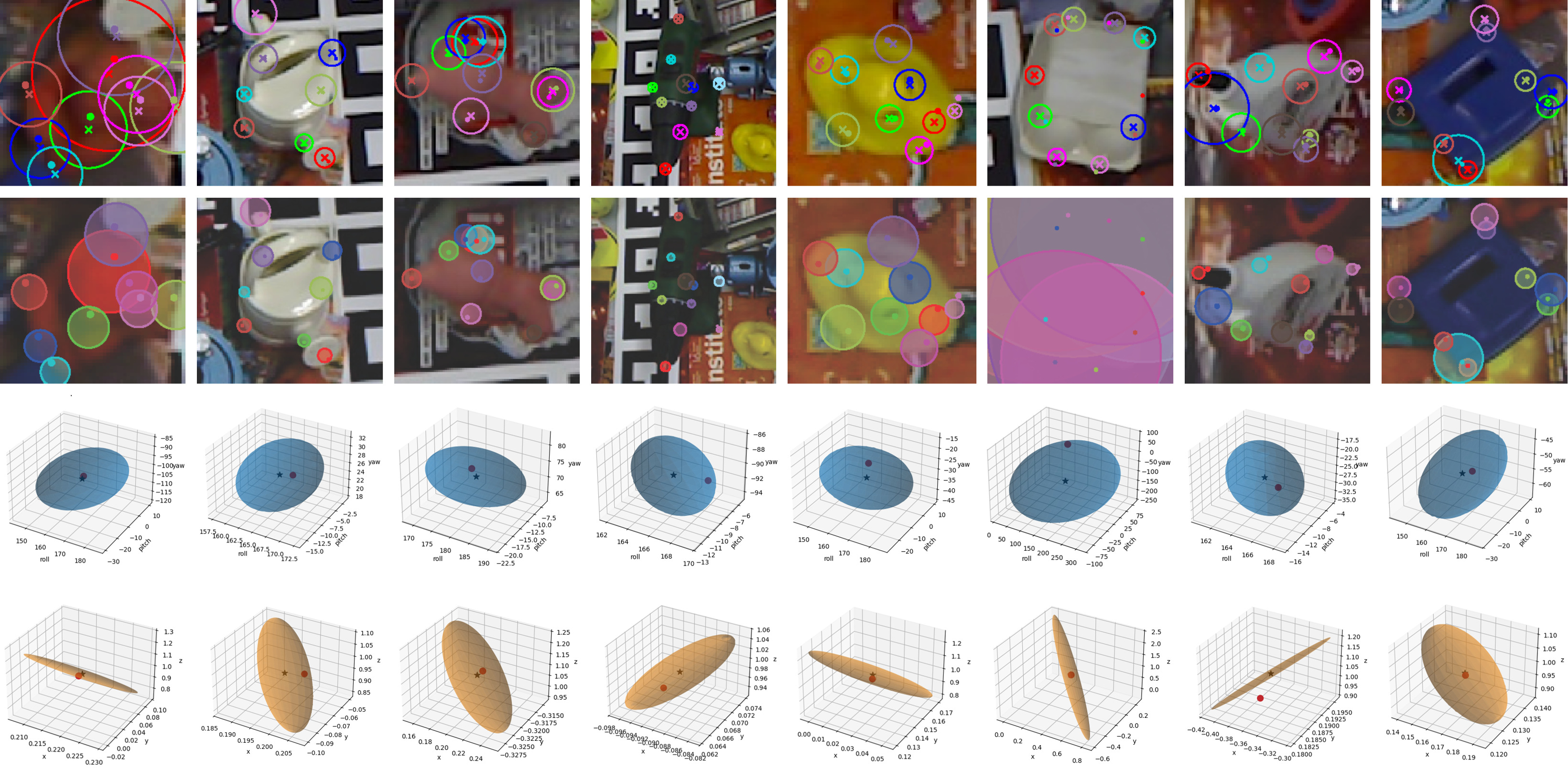}  
    \caption{$\mathcal{P}_{\Theta}(\mathbf x_n|\mathbf{I})$, $\mathcal{K}^{r}$, $\mathcal{R}^{d}_{r}$, and $\mathcal{T}^{d}_r$ (corresponding to rows 1 to 4) for eight objects (corresponding to columns 1 to 8) in some image of the LMO dataset when $\epsilon = 0.1$.}
    \label{fig_eps_10_lmo}
\end{figure*}

\begin{figure*}[h]
    \includegraphics[width=\textwidth]{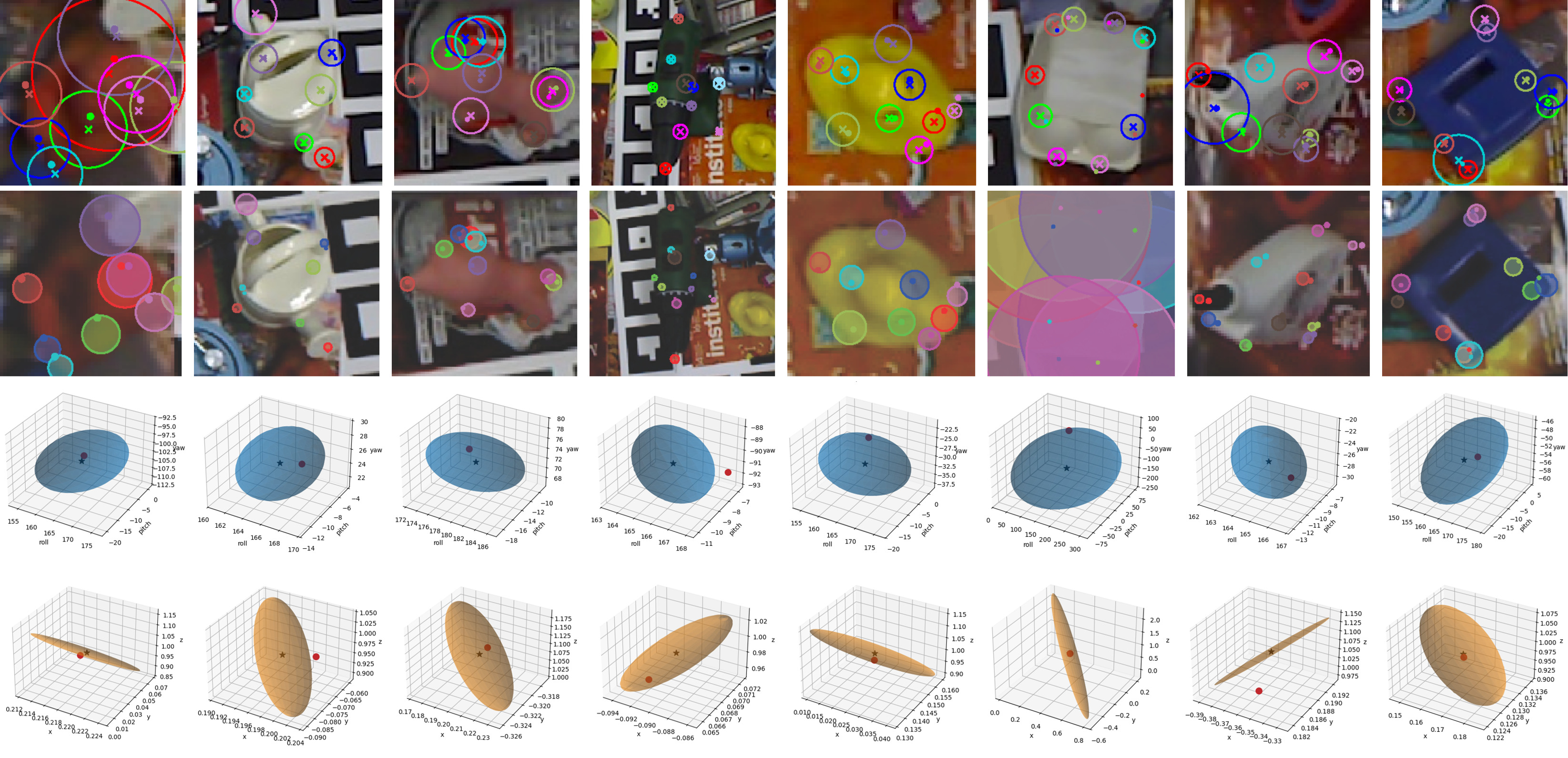}  
    \caption{$\mathcal{P}_{\Theta}(\mathbf x_n|\mathbf{I})$, $\mathcal{K}^{r}$, $\mathcal{R}^{d}_{r}$, and $\mathcal{T}^{d}_r$ (corresponding to rows 1 to 4) for eight objects (corresponding to columns 1 to 8) in some image of the LMO dataset when $\epsilon = 0.4$.}
    \label{fig_eps_40_lmo}
\end{figure*}

Fig.~\ref{fig_eps_10_lmo} and Fig.~\ref{fig_eps_40_lmo} provides 8 LMO objects’ $\mathcal{P}_{\Theta}(\mathbf{x}_n|\mathbf{I})$, $\mathcal{K}^{r}$, $\mathcal{R}^{d}_{r}$, and $\mathcal{T}^{d}_r$ with $\epsilon=0.1$ and $\epsilon=0.4$.
 


\subsection{Single-shot PnP}
\label{sec:single_shot_pnp}

Unlike sampling methods, 
in this paper we follows~\cite{wang2024monocular} utilizing the single-shot PnP weighted by \(\boldsymbol \sigma_n\), thereby significantly enhancing the algorithm's efficiency.
To incorporate the uncertainty into our single-shot PnP framework, we assign weights to \(\tilde{\mathbf x}_{n}\) based on \(\boldsymbol{\sigma}^{n}\). 
The optimization problem for pose estimation can be expressed as follows:

\begin{equation}
    \label{eq_weighted_pnp}
    \min_{\mathbf{R}, \mathbf{t}} \sum_{n=1}^N \rho\left( 
    \mathbf{r}(n)^\top
    \left(\boldsymbol{\sigma}^n\right)^{-1}
    \mathbf{r}(n)
    \right),
\end{equation}

where \(\mathbf{r}(i)\) denotes the reprojection error, defined as:

\begin{equation}
    \label{eq_reproject_error}
    \mathbf{r}(n) = 
    \lambda_i 
    \begin{bmatrix}
        \mathbf{x}^n \\
        1
    \end{bmatrix} - \mathbf{K}\left[\mathbf{R}|\mathbf{t}\right]
    \begin{bmatrix}
        \mathbf{z}^n \\
        1
    \end{bmatrix}.
\end{equation}
In~\cref{eq_reproject_error}, 
\(\lambda_i\) represents the depth value associated with the 3D point \(\mathbf{z}^n\), 
used to scale the projection onto the 2D image plane accurately.
In~\cref{eq_weighted_pnp}, \(\rho(\cdot)\) represents a robust loss function. 
Consistent with the methodology of Wang~\etal~\cite{wang2024monocular}, 
we employ the Huber loss as the robust estimation function, given by:

\begin{equation}\label{eq:huber}
    H(e) = \begin{cases}
        \dfrac{1}{2}e^2, &\text{if } e\le \delta\\
        \delta (|e|-\dfrac{1}{2}\delta), &\text{otherwise}
    \end{cases}.
\end{equation}

Conventional PnP algorithms are prone to inaccuracies from poorly localized keypoints. 
By integrating uncertainty weights into our single-shot PnP approach, we enhance pose estimation accuracy and overcome the limitations of traditional methods.
\end{document}